\documentclass[a4paper,11pt]{article}

\usepackage{amssymb}
\usepackage{amsmath,amsfonts}
\usepackage[linesnumbered,ruled,vlined]{algorithm2e}
\usepackage{algorithmic}
\usepackage{array}
\usepackage{subfigure}
\usepackage{textcomp}
\usepackage{stfloats}
\usepackage{url}
\usepackage{verbatim}
\usepackage{graphicx}
\usepackage{hyperref}
\usepackage{multirow}
\usepackage{makecell}
\usepackage{lscape}
\usepackage{booktabs}
\usepackage[utf8]{inputenc} 
\usepackage[T1]{fontenc}    
\usepackage{amsmath,amssymb,amsfonts} 
\usepackage{hyperref}       
\usepackage{graphicx}       
\usepackage{geometry}       
\title{A Decision-Based Heterogenous Graph Attention Network for Multi-Class Fake News Detection}
\author{%
	Batool Lakzaei\textsuperscript{1}\thanks{Email: \href{mailto:b\_lakzaei@aut.ac.ir}{b\_lakzaei@aut.ac.ir}},
	Mostafa Haghir Chehreghani\textsuperscript{1}\thanks{Corresponding author. Email: \href{mailto:mostafa.chehreghani@aut.ac.ir}{mostafa.chehreghani@aut.ac.ir}} and 
	Alireza Bagheri\textsuperscript{1}\thanks{Email: \href{mailto:ar\_bagheri@aut.ac.ir}{ar\_bagheri@aut.ac.ir}}
}
\date{\footnotesize\textsuperscript{\textbf{1}}Department of Computer Engineering,
	Amirkabir University
	of Technology (Tehran Polytechnic)\\
	Tehran, Iran}

\begin{document}

\maketitle

\begin{abstract}
A promising tool for addressing fake news detection is Graph Neural Networks (GNNs). However, most existing GNN-based methods rely on binary classification, categorizing news as either real or fake. Additionally, traditional GNN models use a static neighborhood for each node, making them susceptible to issues like over-squashing.
In this paper, we introduce a novel model named Decision-based Heterogeneous Graph Attention Network (DHGAT) for fake news detection in a semi-supervised setting. DHGAT effectively addresses the limitations of traditional GNNs by dynamically optimizing and selecting the neighborhood type for each node in every layer. It represents news data as a heterogeneous graph where nodes (news items) are connected by various types of edges.
The architecture of DHGAT consists of a decision network that determines the optimal neighborhood type and a representation network that updates node embeddings based on this selection. As a result, each node learns an optimal and task-specific computational graph, enhancing both the accuracy and efficiency of the fake news detection process.
We evaluate DHGAT on the LIAR dataset, a large and challenging dataset for multi-class fake news detection, which includes news items categorized into six classes. Our results demonstrate that DHGAT outperforms existing methods, improving accuracy by approximately 4\% and showing robustness with limited labeled data.

\end{abstract}

\section{Introduction}
\label{sec:intro}
The rapid expansion of the internet has enabled global connections, with platforms like Facebook, Instagram, and Twitter facilitating communication regardless of location~\cite{hanna2011we}.
These platforms allow users to quickly and efficiently share information, ideas, and opinions with wide audiences~\cite{xie2023heterogeneous}, which has changed how people access information in their daily lives~\cite{cao2021knowledge}.
Today, many prefer online resources and social networks over traditional media for communication, entertainment, and news~\cite{xie2023heterogeneous}.
However, the reliability of information on these platforms is often uncertain~\cite{phan2023fake}, leading to a rise in fake news due to the speed, accessibility, and low cost of social media~\cite{song2021temporally}.
Research shows that fake news spreads faster and wider than real news~\cite{vosoughi2018spread}.
For instance, 50\% of Facebook traffic involves fake news referrals, compared to 20\% from reputable sources~\cite{liao2021integrated}.
The spread of fake news can have harmful effects, such as influencing voter perceptions during elections~\cite{allcott2017social} or causing stock market fluctuations~\cite{ahern2015rumor}.
The challenge of distinguishing real from fake news worsens these effects~\cite{jin2021survey}, making it crucial to develop strategies for automatically detecting and preventing fake news dissemination.

In recent years, researchers have proposed various methods for detecting fake news.
Early approaches used traditional machine learning algorithms like Support Vector Machine (SVM)~\cite{rubin2016fake} and logistic regression~\cite{tacchini2017some}, which, despite their success, relied on manually extracting features from news content, user profiles, and publication patterns~\cite{hu2019multi}.
Given the diversity in topics, styles, and platforms where fake news appears~\cite{shu2017fake}, manually extracting relevant features is time-consuming, costly, and error-prone.
To address this, deep learning methods like LSTM~\cite{rashkin2017truth}, RNN~\cite{ma2016detecting}, and CNN~\cite{zhou2020similarity} were introduced, allowing for automatic feature extraction and improved performance.
However, these methods struggle with short text, a common format on social networks, which remains a significant challenge in detecting fake news~\cite{DBLP:conf/acl/Wang17, hu2019multi}.

Most methods for detecting fake news rely on supervised approaches, but these face challenges due to the scarcity of high-quality labeled datasets, which are difficult and costly to collect and label. Semi-supervised methods offer a potential solution to this issue.
Additionally, many current approaches use binary classification, labeling news as either real or fake.
However, this can miss finer details, as some news may contain partial falsehoods.
Multi-class classification methods provide a more detailed analysis.
For instance, the LIAR dataset~\cite{DBLP:conf/acl/Wang17} categorizes news into six classes, allowing for a more nuanced assessment of news accuracy.

Various features, such as content and context, play a role in detecting fake news, but most methods focus on these separately~\cite{monti2019fake}, assuming that news items are independent~\cite{phan2023fake}.
However, much fake news spreads through social networks, which are graph-based structures where data is interconnected through relationships like citations or interactions~\cite{phan2023fake}.
The choice of data representation is crucial for algorithm performance, especially with unlabeled data~\cite{van2020survey}.
Graph-based models efficiently analyze complex, non-Euclidean data~\cite{breve2011particle} and excel in semi-supervised learning by capturing structural dependencies~\cite{shi2020skeleton}.
Graph Neural Networks (GNNs) have proven effective in handling graph data, using local neighborhood aggregation for better performance~\cite{haghir2022half}.
GNNs have been successfully applied in tasks like object recognition~\cite{shi2020point}, recommender systems~\cite{gao2022graph}, and sentiment analysis~\cite{lu2022aspect}. 
In fake news detection, GNNs outperform other methods by integrating both content and context features in a unified framework~\cite{hu2019multi, ren2020adversarial, benamira2019semi, monti2019fake}.

To address the above challenges, in this paper, we propose a novel semi-supervised Decision-based Heterogeneous Graph Attention Network (DHGAT), that models news data as a heterogeneous graph. This allows each node in each layer to independently choose its optimal neighborhood type, distinct from other nodes and layers.
We evaluate DHGAT on the LIAR dataset, one of the largest and most challenging fake news detection datasets, containing 12,836 mostly short news items classified into six different categories. First, the LIAR dataset is modeled as a heterogeneous graph based on the different features of the news data, with the news items viewed as nodes in the graph. Each node is connected to its neighbors by various types of edges. Each of these neighborhood types represents a different kind of relationship between nodes and has a distinct impact on generating the node embedding vector.
Existing methods in the literature that are based on graphs and operate on the LIAR dataset often process this dataset as a homogeneous graph. On the other hand, most methods in the fake news detection literature that deal with heterogeneous graphs use common techniques such as attention-based methods, like the GAT network, which learns the weight of each neighbor to capture the differences in the impact of various neighborhood types. These networks learn an optimal weight matrix and use it to execute the same message-passing process on all graph nodes in each layer. However, sometimes the necessary and effective neighborhood for each node may differ from other nodes in the same layer and even from its own neighborhood in other layers (for an example and more details, refer to Section~\ref{sec:proposed}).
None of the standard models of graph neural networks or the methods presented in the literature on heterogeneous graphs have the capability to implement such a scenario. These methods lack a mechanism to condition the propagation of information independently for each node in each layer. To implement such a scenario, it is necessary for each node to independently choose its desired neighborhood type in each layer.
Our proposed DHGAT network enables the implementation of such scenarios by leveraging the Gumbel-Softmax distribution, combining and simultaneously training two networks: a decision network (to select the best type of neighborhood) and a representation network (to update the embedding vectors of nodes).
The Gumbel-Softmax distribution is used in the decision network to select the best type of neighborhood in each layer. This distribution provides a continuous approximation of samples from a categorical distribution \cite{DBLP:conf/iclr/JangGP17}, allowing the model to make differentiable choices between different neighborhood types.
The effectiveness of the Gumbel-Softmax distribution has been demonstrated in various GNN tasks, including node clustering \cite{DBLP:journals/corr/abs-2102-10775}, community detection \cite{acharya2020community, chaudhary2023gumbel}, feature selection \cite{acharya2020feature}, and learning message propagation \cite{xiao2021learning, DBLP:conf/icml/FinkelshteinHBC24}. In contexts where the network needs to choose between multiple modes (in our case, the best neighborhood type), the use of Gumbel-Softmax has shown a positive impact on model performance. This makes it well-suited for tasks where adaptive neighborhood selection, as in DHGAT, is essential for improving model accuracy and generalization.
In addition, in the classification layer based on the six classes in the LIAR dataset, we have defined a customized loss function that aims to predict the labels accurately. If the prediction is incorrect, the loss function reduces the semantic distance between the predicted labels and the actual labels.

Our contributions can be summarized as follows:
\begin{itemize}
	\item 
	We introduce a novel approach by utilizing speaker profile information as contextual data to model the LIAR dataset as a heterogeneous graph. In this graph, while the nodes are of the same type, the edges are of different types, capturing various relationships between the news items. Each node in the graph can be connected to others through multiple types of edges, each representing a specific type of relationship between the news items. To the best of our knowledge, this is the first attempt to model the LIAR dataset as a heterogeneous graph using diverse contextual information.
	\item
	Unlike many existing methods that rely on supervised approaches and reduce the LIAR dataset to a binary classification problem, we adopt a challenging semi-supervised and multiclass classification approach. Specifically, we propose a semi-supervised method for six-class classification of fake news in the LIAR dataset, which tackles the complexity inherent in multiclass scenarios.
	\item
	We propose a novel architecture for Graph Neural Networks (GNNs), named DHGAT (Decision-based Heterogeneous Graph Attention Network). In DHGAT, each node independently learns the optimal neighborhood type for updating its embedding vector at each layer, irrespective of other nodes and layers. In the constructed heterogeneous graph, each node can have various types of neighborhoods, consisting of one or a combination of different edge types in the graph. The impact of these neighborhood types varies across different nodes and layers, as different nodes require information from different neighborhood types at each layer. The proposed DHGAT model, which includes two networks--a decision network ($\Phi$) and a representation network ($\Psi$)--allows each node to independently select the best neighborhood type for updating its embedding vector in the current layer.
	\item
	We discuss the properties of the DHGAT model and demonstrate its suitability for long-range applications. Additionally, we show that DHGAT positively impacts mitigating the effects of over-squashing and over-smoothing, which are significant challenges in GNNs.
	\item
	Our empirical evaluation results indicate that the DHGAT model outperforms existing methods even with a limited amount of labeled data (10\%, 20\%, and 30\%), showing superior performance in the task of fake news classification.
\end{itemize}

The remainder of this paper is organized as follows.
Section \ref{sec:rw} reviews the related works in the field, providing a foundation for understanding the context of our research. In \ref{sec:problem_def}, we define the problem that motivates our study.
Section \ref{sec:proposed} presents the proposed model, detailing its design and underlying principles. The properties and theoretical underpinnings of the model are discussed in
Section \ref{sec:properties}.
In Section \ref{sec:exp}, we describe the experiments conducted to evaluate the model's performance, followed by an analysis of the results. Finally, \ref{sec:conclusion} concludes the paper.

\section{Related work}
\label{sec:rw}

There are two main strategies for combating fake news \cite{lakzaei2024disinformation}: intervention and detection. Detection approaches concentrate on identifying and distinguishing fake news from legitimate information. In contrast, intervention strategies seek to limit the spread of fake news by identifying malicious sources \cite{zhang2023finding, wang2024distributed} or fraudulent users \cite{asghari2022using, nivas2024fake}. The method proposed in this paper falls under the detection category, so this section will focus on detection methods.

Different methods for detecting fake news have been extensively explored in the literature, each employing distinct features, models, and approaches to differentiate between fake and real news. The majority of these methods are grounded in a supervised learning approach.
Yadav et al. \cite{yadav2024emotion} introduce FNED, a deep neural network model designed for the rapid identification of fake news on social media. By integrating user profile analysis with textual data analysis techniques like CNNs and attention mechanisms, FNED detects fake news across various content formats.
Qu et al. \cite{qu2024qmfnd} introduce QMFND, a quantum model for fake news detection that combines text and image analysis using quantum convolutional neural networks (QCNN).
The NSEP framework \cite{fang2024nsep} detects fake news by analyzing both macro and micro semantic environments. It first categorizes news into these environments and then uses GCNs and attention mechanisms to identify semantic inconsistencies between news content and associated posts.

One of the most significant challenges in detecting fake news using supervised learning approaches is the need for large and high-quality labeled datasets, which requires substantial time and financial and human resources. 
To address this challenge, researchers have turned to semi-supervised approaches.
DEFD-SSL \cite{al2023robust} is a semi-supervised method for detecting fake news that integrates various deep learning models, data augmentations, and distribution-aware pseudo-labeling. It uses a hybrid loss function that combines ensemble learning with distribution alignment to achieve balanced accuracy, especially in imbalanced datasets.
Shaeri and Katanforoush \cite{shaeri2023semi} introduce a semi-supervised approach for fake news detection, which combines LSTM with self-attention layers. To handle data scarcity, they suggest a pseudo-labeling algorithm and utilized transfer learning from sentiment analysis through pre-trained RoBERTa pipelines to further enhance accuracy.
Canh et al. \cite{canh2023fake} propose MCFC, a fake news detection method using multi-view fuzzy clustering on data from various sources. It extracts features like title and social media engagement, applying fuzzy clustering and semi-supervised learning. MCFC improves accuracy but faces challenges with parameter complexity and computational inefficiency.
WeSTeC \cite{akdag2024early} automates labeling by generating functions from content features or limited labeled data, applying them to unlabeled data with weak labels aggregated via majority vote or probabilistic models. It uses a RoBERTa classifier fine-tuned for the dataset's domain, handling semi-supervision and domain adaptation by incorporating labeled data from various domains.

Despite the fact that multi-class classification of fake news can offer a more accurate and nuanced analysis compared to binary classification, many of the presented methods (both supervised and semi-supervised) still classify news into just two classes:  fake or real, and few researchers have explored multi-class classification in depth.
Pritzkau \cite{pritzkau2021nlytics} present a method for detecting fake news by fine-tuning RoBERTa, a Transformer-based model, for a 4-class classification task.
The approach involves training on concatenated titles and bodies of articles, employing word-level embeddings and multi-head attention for feature processing.
Rezaei et al. \cite{rezaei2022early} propose a 5-class fake news detection model using a stacking ensemble of five classifiers--random forest, SVM, decision tree, LightGBM, and XGBoost--combined with AdaBoost for improved accuracy. The model is trained on PolitiFact data and leverages features like sentiment and semantic analysis.
Majumdar et al. \cite{majumdar2021multi} present a deep learning approach for 4-class fake news detection using LSTM networks. Data preprocessing steps include removing stop words, correcting spelling errors, and eliminating punctuation. 

In this paper, we focus on the LIAR dataset \cite{DBLP:conf/acl/Wang17}, which is one of the largest and most challenging multi-class fake news datasets. Although many researchers have used this dataset to evaluate their proposed methods, most have converted the 6-class dataset into a binary one \cite{pszona2023towards, choudhury2023novel, tsai2023stylometric, li2024re, guo2022mixed, xie2023heterogeneous, cui2023intra}. This conversion simplifies the classification task but may overlook the nuances of multi-class classification. Consequently, only a few methods have been proposed for multi-class classification of fake news using this dataset \cite{hu2019multi, liao2021integrated, goldani2021detecting, balshetwar2023fake, goldani2021convolutional}.
In addition, many methods applied to the LIAR dataset rely on text processing approaches. The textual content in this dataset is often very short, which can be challenging for methods that focus solely on text due to the lack of context and details. Therefore, incorporating hybrid approaches that employ additional features, such as speaker profiles, alongside textual content can enhance the effectiveness of classification by providing more comprehensive information.
Goldani et al. \cite{goldani2021convolutional} present a CNN-based model for fake news detection, which uses margin loss to enhance feature discrimination. The model incorporates pre-trained word embeddings like GloVe and explores static, non-static, and multichannel word2vec models. The CNN extracts features using convolutional filters, while fully connected layers perform classification with the aid of margin loss to improve accuracy.
The NER-SA model \cite{tsai2023stylometric} combines Named Entity Recognition (NER) with Natural Language Processing (NLP) to enhance fake news detection. It creates entity banks from related and unrelated news to differentiate genuine from fake news. By applying a mathematical model and a simulated annealing algorithm, the model defines a legitimate area, based on named entity patterns. 
The proposed method in \cite{balshetwar2023fake} uses sentiment analysis with a lexicon-based scoring algorithm to identify key feature words and calculates their propensity scores. Additionally, it employs a multiple imputation strategy (MICE) to address multivariate missing data, and applies Term Frequency-Inverse Document Frequency (TF-IDF) to extract effective features from news content. Various classifiers, including Naive Bayes, passive-aggressive, and deep neural networks are used to evaluate the model's performance.

After recognizing the potential of Graph Neural Networks (GNNs) in various tasks, including fake news detection, some researchers have attempted to apply these networks to the LIAR dataset. The use of GNNs allows for the simultaneous analysis of both content and context features of the data, which can significantly enhance the performance of fake news detection models.
Hu et al. \cite{hu2019multi} introduce Multi-depth Graph Convolutional Networks (M-GCN) for fake news detection on the LIAR dataset. M-GCN enhances detection by utilizing graph embeddings and multi-depth GCN blocks to capture multi-scale information from news node neighbors, which is then integrated through an attention mechanism.
Guo et al. \cite{guo2022mixed} present FRD-VSN, a fake content detection model designed for vehicular social networks (VSN). This model combines GNNs with CNNs and Recurrent Neural Networks (RNN) to enhance detection accuracy. By integrating local and global semantic features, the model improves the detection of fake news.
Liao et al. \cite{liao2021integrated} introduce a multi-task learning model, FDML, for fake news detection and topic classification, with a particular focus on improving detection for short news content. The model is based on two key observations: certain topics and authors are more likely to produce fake news. FDML simultaneously trains for both fake news detection and topic classification using a news graph (N-Graph) to integrate textual and contextual information, and employs dynamic weighting to balance the tasks.
Heterogeneous Graph Neural Network via Knowledge Relations for Fake News Detection (HGNNR4FD) \cite{xie2023heterogeneous} builds a heterogeneous graph (HG) from news, entities, and topics and incorporates relations from Knowledge Graphs (KGs). The framework uses attention mechanisms to aggregate information from HGs and KGs, to generate news embeddings that enhance the performance of the fake news detector.
Cui et al. \cite{cui2023intra} present a model, called Intra-graph and Inter-graph Joint Information Propagation Network (IIJIPN), with a Third-order Text Graph Tensor (TTGT) to detect fake news by utilizing sequential, syntactic, and semantic features from text. It addresses challenges like data imbalance and sparse features using data augmentation techniques. TTGT captures contextual relationships within news content, while the IIJIPN framework allows information to propagate both within individual graphs and between graphs, by combining both homogeneous and heterogeneous information. Attention mechanisms are used to generate news representations for classification.
Lakzaei et al. \cite{lakzaei2024loss} propose a semi-supervised, one-class approach for fake news detection, called LOSS-GAT. They utilize a two-step label propagation method with GNNs to initially classify news as fake or real. The graph structure is further refined through structural augmentation techniques, and final labels for unlabeled data are predicted using a GNN with a randomness-inducing aggregation function in local node neighborhoods.

To address the challenges of existing methods (supervised learning, binary classification of fake news, and reliance on textual content without contextual information), we propose the Decision-based Heterogeneous Graph Attention Network (DHGAT). This is a multi-class semi-supervised classification method based on a graph neural network applied to a heterogeneous graph for the LIAR dataset, incorporating both textual and non-textual features to enhance classification performance.
\autoref{tbl: liar_methods} compares our proposed method with several existing approaches on the LIAR dataset.

\begin{table}[]
	\centering
	\tiny
	\caption{Overview of different approaches for fake news detection on the LIAR dataset.\label{tbl: liar_methods}}
	\resizebox{\textwidth}{!}{%
		\begin{tabular}{@{}cccccccc@{}}
			\toprule
			Method &
			\textbf{\makecell{semi/supervised\\learning}} &
			\textbf{\makecell{binary/multi-class\\classification}} &
			\textbf{\makecell{textual\\features}} &
			\textbf{\makecell{non-textual\\features}} &
			\textbf{\makecell{graph\\modeling}} &
			\textbf{\makecell{homogeneous/\\heterogeneous}} &
			\textbf{GNN} \\ \midrule
			\cite{pszona2023towards}         & supervised      & binary      & \checkmark &   &   &              &   \\ \midrule
			\cite{choudhury2023novel}         & supervised      & binary      & \checkmark &   &   &              &   \\ \midrule
			\cite{tsai2023stylometric}         & supervised      & binary      & \checkmark &   &   &              &   \\ \midrule
			\cite{li2024re}         & supervised      & binary      & \checkmark &   &   &              &   \\ \midrule
			\cite{balshetwar2023fake}         & supervised      & multi-class & \checkmark &   &   &              &   \\ \midrule
			\cite{goldani2021detecting}         & supervised      & multi-class & \checkmark & \checkmark &   &              &   \\ \midrule
			\cite{goldani2021convolutional}         & supervised      & multi-class & \checkmark & \checkmark &   &              &   \\ \midrule
			\cite{guo2022mixed}         & supervised      & binary      & \checkmark & \checkmark & \checkmark & homogeneous  & \checkmark \\ \midrule
			\cite{hu2019multi}         & semi-supervised & multi-class & \checkmark & \checkmark & \checkmark & homogeneous  & \checkmark \\ \midrule
			\cite{liao2021integrated}          & supervised      & multi-class & \checkmark & \checkmark & \checkmark & homogeneous  & \checkmark \\ \midrule
			\cite{cui2023intra}         & supervised      & binary      & \checkmark & \checkmark & \checkmark & heterogenous & \checkmark \\ \midrule
			\cite{xie2023heterogeneous}          & semi-supervised & binary      & \checkmark & \checkmark & \checkmark & heterogenous & \checkmark \\ \midrule
			DHGAT(ours) & semi-supervised & multi-class & \checkmark & \checkmark & \checkmark & heterogenous & \checkmark \\ \bottomrule
		\end{tabular}%
	}
\end{table}

\section{Problem definition}
\label{sec:problem_def}

Let $D = \{d_1, d_2, \ldots, d_n \}$ be the news set with $n$ news items, where each news item $d_i$ contains text content $t_i$ and a contextual information set $S_i$. The contextual information $S_i$ is drawn from a set $S = \{s_1, s_2, \dots, s_m\}$,
representing $m$ types of side information available in the dataset.
We formulate the problem of fake news detection as a multi-class classification task utilizing a semi-supervised approach. The class labels are represented by $y \in \{y_1, y_2, \ldots, y_C\}$, where $C$ indicates the number of classes. For instance, in the LIAR dataset, which we focus on in this study,
$C = 6$.
To approach this problem, we first construct a heterogeneous graph $HG(V, E, R)$ using the contextual information, where $V$ is the set of nodes, $E$ is the set of edges, and $R$ is a set of edge types. In this graph, $|V| = n$, meaning that each news item is represented as a node, and $E = \{(v_i, v_j, e_{type} = r) \vert v_i \in V, v_j \in V, r \in R\}$.
Thus, all nodes are of the same type, but the edges can vary in type, represented by the set $R$.
The edge types are determined based on the contextual information $S$, implying $|R| \leq |S|$.
Different subsets or all of the contextual features can be selected to define various edge types in the graph.

Following the semi-supervised approach, labels for a small subset of news, denoted as $D_L \subset D$, are observed. Our goal is to learn the model $F$ for predicting labels for all unlabeled data $D_U = D - D_L$, utilizing the labeled news:
\begin{equation}
	\hat{y} = F(d_i),
\end{equation}
where $\hat{y} \in \{y_1, y_2, \ldots, y_C\} $ is the predicted label, and $d_i \in D_U$ is an unlabeled example.

We summarize frequently used notations in \autoref{tab:notations}
\begin{table}[!t]
	\centering
	\caption{Notations and their descriptions.\label{tab:notations}}	
	\small
	\begin{tabular}{l l}
		\hline
		\textbf{Notation} & \textbf{Definition} \\ \hline
		
		$D = \{d_1, d_2, \dots , d_n\}$		& set of news items \\ \hline
		
		$n$		& number of news items in $D$ \\ \hline		
		
		$D_L \subset D$		& set of labeled news items \\ \hline
		
		$D_U \subset D$		& set of unlabeled news items \\ \hline
		
		$y \in \{y_1, y_2, \ldots, y_C\} $		& ground-truth class label \\ \hline
		
		$\hat{y} \in \{y_1, y_2, \ldots, y_C\}$		& predicted class label \\ \hline
		
		$C$ & number of classes \\ \hline
		
		$X \in \Bbb{R}^{n \times l_t}$		& textual feature vectors \\ \hline
		
		$l_t$		& length of textual feature vectors \\ \hline
		
		$S = \{s_1, s_2, \dots, s_m\}$ & set of side information available in the dataset \\ \hline
		
		$V = \{v_1 , d_2, \dots , d_n\}$ & set of vertices \\ \hline
		
		$E = \{(v_i, v_j, e_{type} = r) \vert v_i , v_j \in V, r \in R \}$ & set of edges \\ \hline
		
		$R = \{r_1, r_2, \dots, r_m\}$ & set of edge types \\ \hline
		
		$HG(V,E,R)$ & heterogeneous graph \\ \hline
		
		$\Gamma = \{ \gamma_0 , \gamma_2 , \dots , \gamma_{\vert \Gamma \vert -1} \} $ & set of neighborhood types \\ \hline
		
		$\rho^l_v \in \Bbb{R}^{\vert \Gamma \vert}$ & \makecell[l]{probability distribution over possible \\ neighborhood types $\Gamma$ for node $v$ } \\ \hline
		
		$N_d(v) = \{u \vert (v,u, e_{type}=d) \in E , d \in \Gamma\}$ & neighbors of node $v$ with edge type $d$ \\ \hline
		
		$h^l_v \in \Bbb{R}^{d_l}$ & embedding of node $v$ in layer $l$ \\ \hline
		
		$d_l$ & size of embeddings in layer $l$ \\ \hline		
	\end{tabular}
\end{table}


\section{Proposed method}
\label{sec:proposed}

\begin{figure}[!t]
	\centering
	\includegraphics[scale=.75]{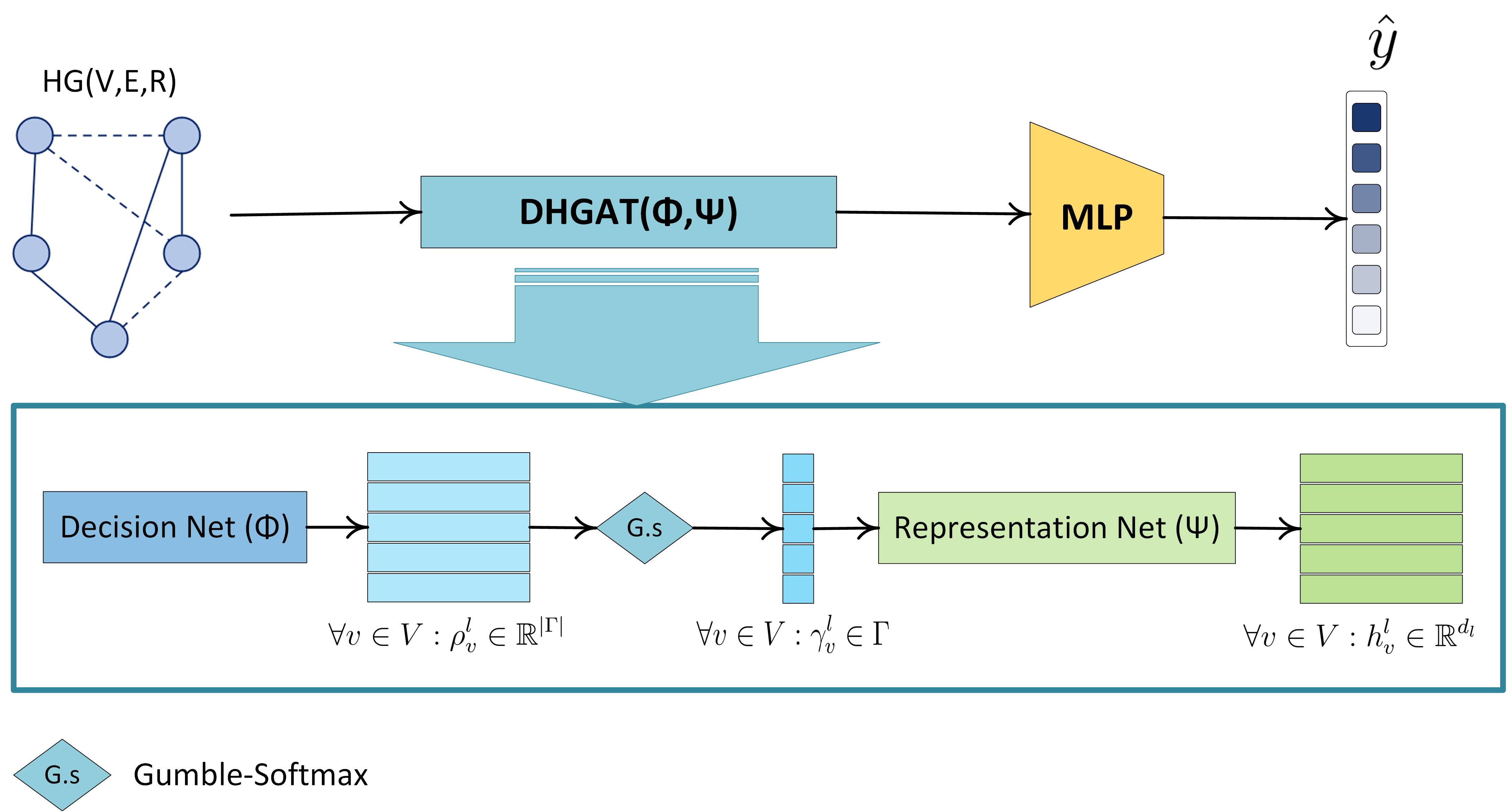}
	\caption{An overview of the proposed model DHGAT.	
		It takes a heterogeneous graph $HG(V, E, R)$ with various types of edges as input.
		Initially, using the decision network $\Phi$, which is based on GATv2 \cite{DBLP:conf/iclr/Brody0Y22}, a probability vector $\rho$ is generated for each node at every layer, where each component represents the probability of a particular neighborhood type $\gamma_i$.
		Then, the Gumbel-Softmax estimator is used to determine a specific neighborhood type.
		In the subsequent step, the representation network $\Psi$, also based on GATv2, updates each node's embedding vector based on the determined neighborhood type in the decision network.
		Finally, the generated embedding vectors are sent to the classification layer, where the final label for each node is predicted.\label{fig:dhgat}}	
\end{figure}

In this paper, we propose a Decision-based Heterogeneous Graph Attention Network (DHGAT) for detecting fake news using a semi-supervised approach. 
Figure~\ref{fig:dhgat} illustrates the structure of our proposed model, which includes three main components: 1) heterogeneous graph construction, 2) Decision-based Heterogeneous Graph Attention Network (DHGAT), and 3) multi-class classification.

We use the FastText \cite{bojanowski2017enriching} model to generate textual embeddings for the news.
FastText improves upon Word2Vec \cite{DBLP:journals/corr/abs-1301-3781} by incorporating subword information, representing each word as a collection of character n-grams. 
This approach enables FastText to handle rare words, misspellings, and morphologically rich languages more effectively, generating meaningful word vectors even for unseen words.
After generating text embeddings, we construct a heterogeneous graph comprising nodes of the same type but connected by different edge types, representing various relationships between the news items.
The decision-based deterogeneous graph attention network processes this graph, where each node in each layer independently selects the most effective type of neighborhood to update its embedding vector.
The final embedding vectors are fed into a Multi-Layer Perceptron (MLP) classifier, which uses a custom loss function to not only predict the labels but also minimize semantic errors between predicted and ground-truth labels. 
Detailed explanations of these components are provided in the subsequent sections.

\subsection{Heterogeneous graph construction}
\label{sec:hetero_graph}

The LIAR dataset contains a number of speaker profiles, such as speaker name, party affiliation, job title, and more.
We utilize these features to construct a heterogeneous graph with uniform node types but different edge types.
Each news item is considered a node.
Then, two or more of these speaker profile features are selected, and for each one, a different type of edge is created between the nodes.
For example, if "job title" and "speaker" are chosen, a heterogeneous graph $HG(V,E,R)$
is constructed as follows:
\begin{equation}
	\label{eq:HG}
	HG(V , E , R) =
	\begin{cases}
		R = \{ \text{'job-title'}, \text{'speaker'} \} \\
		V = D  \text{ and  }  |V| = n \\
		E \subseteq \{(v_i, v_j, e_{type} = r) \vert
		v_i , v_j \in V, r \in R \} \\
	\end{cases}
\end{equation}
An edge of type $r$ is created between two nodes if the corresponding news items for these nodes in the dataset share the same value for the feature corresponding to $r$. In other words:
\begin{equation}
	\label{eq:E}
	(v_i, v_j, e_{type} = r) \in E : 
	\begin{cases}
		r = \text{'job-title'  and  $v_i$ and $v_j$ have the same job title} \\
		r = \text{'speaker'  and  $v_i$ and $v_j$ have the same speaker}
	\end{cases}
\end{equation}

\subsection{Decision-based Heterogeneous Graph Attention Network (DHGAT)}
\label{sec:DHGAT}

\begin{figure}[!t]
	\centering
	\includegraphics[scale=.55]{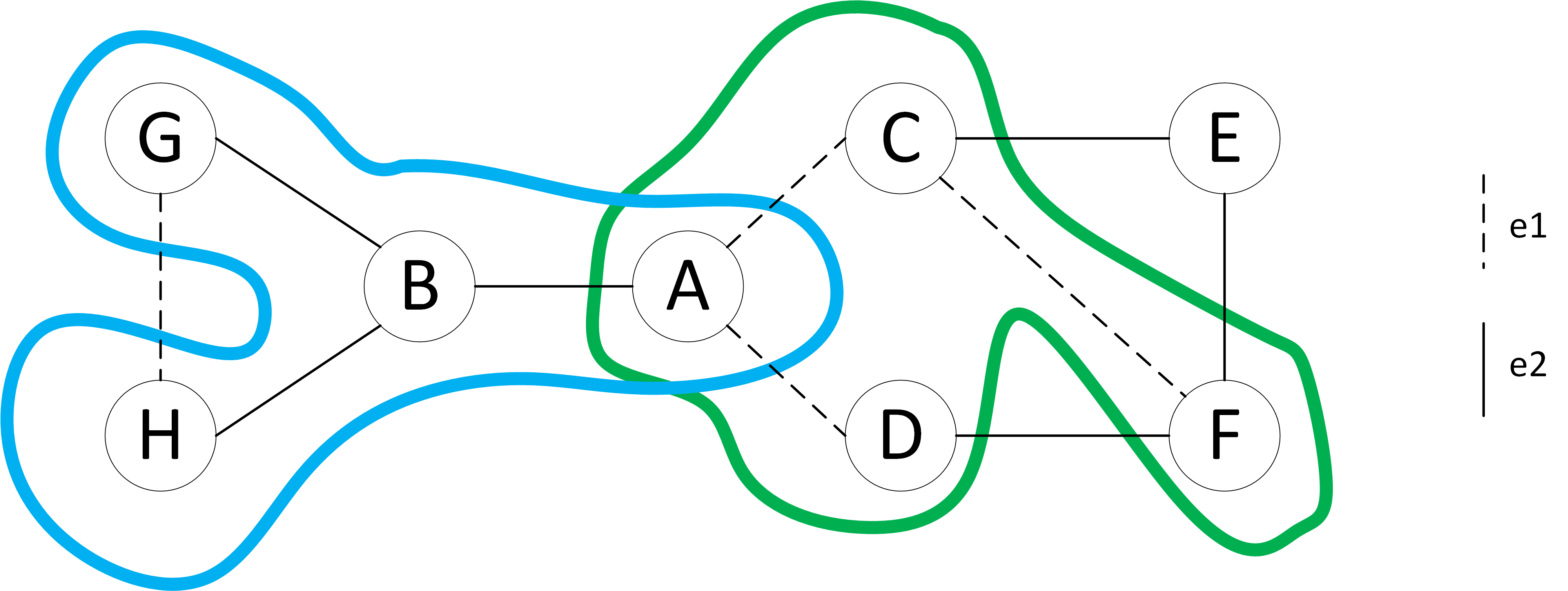}
	\caption{Illustration of dynamic neighborhood selection in a heterogeneous graph :
		(a) In Scenario 1, nodes $A$ and $B$ independently select different edge types ($ e_1$ and $e_2$, respectively) to update their embeddings.
		(b) In Scenario 2, node $A$ selects different edge types across three layers to optimally update its embedding.\label{fig:sample}}	
\end{figure}

In the heterogeneous graph $HG$, each node can connect to neighbors through various edge types, representing different relationships between news items.
This results in a diverse set of neighborhood types for each node, with each type potentially transmitting unique information.
These varying neighborhood types can differently influence the generation of the node's embedding.
While existing GNNs for heterogeneous graphs use methods like multi-level attention mechanisms \cite{wang2019heterogeneous, hu2020heterogeneous} or distinct weight matrices \cite{fu2020magnn} to capture and differentiate the effects of these neighborhood types, they overlook an important aspect: the need for different nodes to have different neighborhood requirements.
This means that two nodes $v$ and $u$ in the same layer might need different types of neighborhoods, or a single node $v$ might require different neighborhood types across different layers.
Current GNNs cannot accommodate this scenario because they lack the ability to condition the message-passing process.
Instead, it is preferable to allow each node, in each layer, to independently decide which type of neighborhood to use for updating its own embedding vector.
In other words, it is necessary for the message-passing process across layers to be dynamic, allowing each node to adaptively select the most relevant type of neighborhood as the optimal neighborhood type for updating its embedding at each layer.
As a motivating example, consider the heterogeneous graph with two edge types, $e_1$ and $e_2$, depicted in \autoref{fig:sample}, and examine the following two scenarios:
\begin{itemize}
	\item \textbf{example scenario 1:}
	suppose that at layer l, node $A$ requires information only from neighbors connected by edges of type $e1$, while node $B$ needs information only from neighbors connected by edges of type $e2$.
	In this case, node $A$ should update its embedding using only the information from a subgraph highlighted in green, 
	and node $B$ should update its embedding using only the information from a subgraph highlighted in blue.
	
	\item \textbf{example scenario 2:}
	suppose that in a 3-layer GNN, node $A$ requires  different information in different layers:
	\begin{itemize}
		\item in layer 1, it needs information only from neighbors connected by edges of type $e_1$: $\{C, D\}$.
		\item in layer 2, it needs information only from neighbors connected by edges of type $e_2$: $\{B\}$.
		\item in layer 3 it needs information from both edge types: $\{B, C, D\}$.
	\end{itemize}
\end{itemize}

In this section, we introduce DHGAT, which enables each node in each layer to independently learn which type of neighborhood to use for updating its own embedding.
This decision is made independently of the choices made by other nodes in the same layer and the choices made by the same node in other layers. 
The number of possible choices will vary depending on the diversity of neighborhoods. 
In general, if the number of edge types in the heterogeneous graph is $\vert R \vert$, then a maximum of $\vert \Gamma \vert = 2^{\vert R \vert}$ different neighborhood types can be defined:
\begin{equation}
	\label{eq:neigh-types}
	\Gamma = \{ \gamma_0 , \gamma_2 , \dots , \gamma_{\vert \Gamma \vert -1} \} , \gamma_i \subset R
\end{equation}
Based on the above equation, we can define the set of edges for each type of neighborhood as follows:
\begin{equation}
	\label{eq:E-types}
	E_{\gamma_i} = \{ (v,u, e_{type}) \in E \vert
	e_{type} \in {\gamma_i}\}
\end{equation}

For example, if HG is the heterogeneous graph defined in \autoref{eq:HG}, there are two types of edges:
$ e_{type} \in R =  \{ \text{'job-title'} , \text{'speaker'} \}$.
Therefore, each node $v$ can choose from 4 types of neighborhoods in each layer:
\begin{itemize}
	\item \textbf{none:} $v$ does not use any neighbors and behaves as an isolated node:
	\begin{equation}
		\begin{cases}
			\gamma_0 = \{\} \\
			E_{\gamma_0} = \{\}
		\end{cases}
	\end{equation}
	
	\item \textbf{job-title:} $v$ uses neighborhood with $e_{type} = \text{'job-title'}$:
	\begin{equation}
		\begin{cases}
			\gamma_1 = \{ \text{'job-title'} \} \\
			E_{\gamma_1} = \{ (v,u, e_{type}) \in E \vert
			e_{type} = \text{'job-title'}\}
		\end{cases}
	\end{equation}

	\item \textbf{speaker:} $v$ uses neighborhood with $e_{type} = \text{'speaker'}$:
	\begin{equation}
		\begin{cases}
			\gamma_2 = \{ \text{'speaker'} \} \\
			E_{\gamma_2} = \{ (v,u, e_{type}) \in E \vert
			e_{type} = \text{'speaker'}\}
		\end{cases}
	\end{equation}
	
	\item \textbf{both:} $v$ uses both types of neighborhoods:
	\begin{equation}
		\begin{cases}
			\gamma_3 = \{ \text{'job-title'} ,  \text{'speaker'} \} \\
			E_{\gamma_3} = \{ (v,u, e_{type}) \in E \vert e_{type} = \text{'job-title'} \lor
			e_{type} = \text{'speaker'}\}
		\end{cases}
	\end{equation}
\end{itemize}
Our proposed network, DHGAT, allows each node $v$ to independently select one of the above modes that provides the best information for updating its embedding.

In DHGAT, we need to predict categorical neighborhood types for each node in each layer.
It poses a challenge for gradient-based optimization due to its non-differentiable nature.
The Gumbel-Softmax estimator \cite{DBLP:conf/iclr/JangGP17, DBLP:conf/iclr/MaddisonMT17, DBLP:conf/icml/FinkelshteinHBC24} addresses this challenge by providing a differentiable and continuous approximation of discrete sampling.
There is a finite set $\Gamma$ of neighborhood types (\autoref{eq:neigh-types}).
We aim to learn a categorical distribution over $\Gamma$, represented by a probability vector $\rho \in \mathbb{R}^{|\Gamma|}$, where each element stores the probability of a neighborhood type $\gamma_i \in \Gamma$.
Let $\rho(\gamma_i)$ denote the probability of neighborhood type $\gamma_i \in \Gamma$ and $e$ is a one-hot vector representing the selected neighborhood type.
The Gumbel method is a simple and effective approach for sampling $e$ from a categorical distribution:
\begin{equation}
	\label{eq:gumblel}
	e = \text{one-hot}\left(\underset{\gamma_i \in \Gamma}{\text{arg max}}\left(g(\gamma_i) + \log(\rho(\gamma_i))\right)\right)
\end{equation}
where $g(\gamma_i) \sim \text{Gumbel}(0, 1)$ is an independent and identically distributed sample for neighborhood type $\gamma_i$.
The $\text{Gumbel}(0, 1)$ distribution can be sampled using inverse transform sampling by drawing $u_i \sim uniform(0,1)$ 
and computing $g(\gamma_i) = - \log(- \log(u_i))$ \cite{DBLP:conf/iclr/JangGP17}.

The Gumbel-Softmax estimator uses the softmax function as a continuous, differentiable approximation to arg max.
So, the Gumbel-Softmax scores are computed as \cite{DBLP:conf/icml/FinkelshteinHBC24}:
\begin{equation}
	\label{eq:gumble-sm}
	\text{Gumbel-softmax} (\rho; \iota) =
	\frac{exp((log(\rho) + g)/\iota)}
	{\sum_{\gamma_i \in \Gamma} exp((log(\rho(\gamma_i)) + g(\gamma_i))/\iota)},	
\end{equation}
where $\rho \in \mathbb{R}^{|\Gamma|}$ is a categorical distribution, 
$g \in \mathbb{R}^{|\Gamma|}$ is a Gumbel distributed vector,
and $\iota$ is a temperature parameter.
As the temperature $\iota$ decreases, the resulting vector approaches a one-hot vector.
The Straight-Through Gumbel-Softmax estimator uses the Gumbel-Softmax approximation during the backward pass for differentiable updates, while during the forward pass, it uses ordinary sampling.

Formally, the DHGAT($\Phi, \Psi$) architecture consists of two Graph Attention Networks (GATs): a decision network $\Phi$ that selects the optimal neighborhood type, and a representation network $\Psi$ that updates embeddings based on the selected neighborhood type.
Given a heterogeneous graph $HG$ as defined in \autoref{eq:HG}, the DHGAT network updates the embedding $h_v^l$ of each node $v$ in layer $l$ as follows:
First, the decision network $\Phi$ predicts a probability distribution $\rho^l_v \in \mathbb{R}^{|\Gamma|}$ over possible neighborhood types $\Gamma$ for node $v$ 
(for \(HG\) defined in \autoref{eq:HG}, $\Gamma = \{\text{none}, \text{job-title}, \text{speaker}, \text{both}\}$ and $|\Gamma| = 4$):
\begin{equation}
	\label{eq:p_l}
	\rho^l_v = GATv2\left(h^{l-1}_v , \{h^{l-1}_u | u \in N_d(v)\}\right),
\end{equation}
\begin{equation}
	\label{eq:N_d}
	N_d(v) = \{u \vert (v,u, e_{type}=d) \in E , d \in \Gamma\},
\end{equation}
where $\text{GATv2}$ is a GATv2 network \cite{DBLP:conf/iclr/Brody0Y22}, and $N_d(v)$ denotes the set of neighbors considered for decision making.
This set can be defined by the user and may include all or just a subset of neighborhood types. 
The embedding vectors in GATv2 are computed using \cite{DBLP:conf/iclr/Brody0Y22}:
\begin{equation}
	\label{eq:gat_1}
	z_v^l = GATv2\left(z^{l-1}_v , \{z^{l-1}_u | u \in N(v)\}\right) = \parallel_{k=1}^{K} \sigma \left(\sum_{u \in N(v)} \alpha_{vu}^k W^{l-1} z_u^{l-1} \right)
\end{equation}
and
\begin{equation}
	\label{eq:gat_2}
	\alpha_{vu} = a \, \sigma \left(W^{l-1} z_v^{l-1} , W^{l-1} z_u^{l-1} \right),
\end{equation}
where $z_v^l$ represents the embedding of node $v$ in layer $l$, $\parallel$ denotes concatenation, $K$ is the number of attention heads, $\sigma$ is a non-linear activation function,
$N(v)$ is the set of neighbors of $v$,
and $\alpha_{vu}$ is the attention coefficient between $v$ and $u$.
The weight matrix $W$ and parameter $a$ are learned during training.

Next, a neighborhood type $\gamma^l_v$ is sampled from $\rho^l_v$ using the Straight-Through Gumbel-Softmax estimator.
The representation network $\Psi$ then updates embeddings based on the selected neighborhood type for each node:
\begin{equation}
	\label{eq:h}
	h^l_v = GATv2\left(h^{l-1}_v , \{h^{l-1}_u | u \in N_r(v)\}\right),
\end{equation}
\begin{equation}
	\label{eq:N_r}
	N_r(v) = \left\{u \vert (v,u, e_{type}=r) \in E , r=\gamma^l_v \right\},
\end{equation}
where $N_r(v)$ is all nieghbors of node $v$ that are linked to $v$ thorough an edge with type $\gamma^l_v$. 
For example, for HG defined in \autoref{eq:HG}, we have
$\gamma^l_v \in \{\text{none}, \text{job-title}, \text{speaker}, \text{both}\}$.

In our implementation, we utilize the GATv2 network for both the decision and representation networks. However, our framework is flexible and can incorporate any desired GNN architecture.

Details about the training process of the DHGAT network are provided in \autoref{alg:DHGAT}. This algorithm takes a heterogeneous graph $HG$ and the desired neighborhood type $d$ for the decision network as inputs.
In each training iteration, the decision network $\Phi$, which is a GATv2 network, uses the embedding vectors generated in the previous layer for each node and the neighbors within the specified neighborhood type $d$ to produce a vector of $\vert \Gamma \vert$ dimensions for each node. This vector represents the probability of each neighborhood type $\gamma_i \in \Gamma$ for that node. Then, using Gumbel-Softmax, the optimal neighborhood type for each node is determined.
In the next step, the representation network $\Psi$, which in this paper is defined as a GATv2, updates the embedding vector of each node based on the node's previous embedding and the optimal neighbors determined by the decision network $\Phi$.
During the training process, the decision network $\Phi$ learns to identify the best neighborhood type for each node, while the representation network $\Psi$ learns the optimal weights to update the node embeddings based on each node's optimal neighborhood type.

\begin{algorithm}[bht!]
	\small
	\caption{Training process of DHGAT.\label{alg:DHGAT}}
	\SetKwInput{KwInput}{Input}                
	\SetKwInput{KwOutput}{Output}              
	\DontPrintSemicolon
	
	\KwInput{\\		
		$HG = (V,E,R)$: heterogeneous graph  \\		
		$X \in \Bbb{R}^{n \times l_t}$: text embeddings vectors \\
		$max\_epoch$ : number of epochs \\
		$L$: number of GNN layers \\
		$d \in \Gamma$: neighborhood type for decision network $\Phi$					
	}
	\BlankLine	
	\KwOutput{Trained model $DHGAT(\Phi, \Psi)$}
	\BlankLine
	\BlankLine
	\BlankLine
	\ForEach{$v \in V$}
	{	
		$N_d(v) = \{u \vert (v,u, e_{type}=d) \in E , d \in \Gamma\}$ \tcp{\autoref{eq:N_d}}
	}
	
	\BlankLine
	\BlankLine
	
	\tcp{Train DHGAT on HG}
	\ForEach{$epoch : 1..max\_epoch$}
	{	\ForEach{$v \in V$}
		{	
			$h^0_v = X_v$
		}
		\ForEach{$l: 1..L$}
		{	\ForEach{$v \in V$}
			{
				\tcp{Learn the optimal neighborhood type using decision network $\Phi$}
				
				$\rho^l_v = GATv2(h^{l-1}_v , \{h^{l-1}_u | u \in N_d(v)\})$ \tcp{\autoref{eq:p_l}}	
				$\gamma^l_v = \text{Gumble-Softmax}(\rho^l_v)$	\\
				
				\BlankLine
				\BlankLine
				
				\tcp{Update embeddings using representation network $\Psi$}
				
				$N_r(v) = \{u \vert (v,u, e_{type}=r) \in E , r=\gamma^l_v$ \tcp{\autoref{eq:N_r}}
				$h^l_v = GATv2(h^{l-1}_v , \{h^{l-1}_u | u \in N_r(v)\})$ \tcp{\autoref{eq:h}}									
			}
		}
		\BlankLine
		$H^L = [h^L_1, \dots , h^L_n]^T$ \tcp{$n = |V|$}
		$\hat{Y} = MLP(H^L)$	\\
		$loss \leftarrow \autoref{eq:loss}$ \\
		$loss.backward()$			
	}
\end{algorithm}

\subsection{Multi-class classification}
\label{sec:classification}

To classify news, the embeddings $H^L$ generated by DHGAT in the final layer (Layer $L$) are fed into a Multi-Layer Perceptron (MLP) network.
To enhance classification performance, we propose a customized loss function defined as:
\begin{equation}
	\label{eq:loss}
	\mathcal{L}(\Theta) = 
	\lambda_1 
	\left( - \frac{1}{N_L} \sum_{i=1}^{N_L} \sum_{c=1}^{C} y_{i,c} \log(\hat{y}_{i,c}) \right)
	+ 
	\lambda_2
	\left( \frac{1}{N_L} \sum_{i=1}^{N_L} |y_i - \hat{y}_i| \right),
\end{equation}
where 
$N_L = |D_L|$ is the number of labeled samples, $C$ is the number of classes, $y_{i,c}$ is a binary indicator ($0$ or $1$) if class label $c$ is the correct classification for sample
$i$, $\hat{y}_{i,c}$ is the predicted probability that sample $i$ belongs to class $c$, $y_i$ is the true class label for sample $i$, and $\hat{y_i}$ is the predicted label for sample $i$.
The parameters $\lambda_1$ and $\lambda_2$ control the contribution of each term in the loss function. The left term in the loss function aims to predict the true labels for unlabeled samples, while the right term is designed to decrease the semantic distance between the predicted labels and the ground-truth labels.

In the LIAR dataset, news items are classified into six categories: \{false: 0, half-true: 1, mostly-true: 2, true: 3, barely-true: 4, pants-fire: 5\}.
We relabel the dataset to sort these labels from "pants-fire" to "true" with numeric values from 0 to 5 as: \{pants-fire: 0, false: 1, barely-true: 2, half-true: 3, mostly-true: 4, true: 5\}.
This relabeling treats the labels as scores indicating the accuracy of news, where a score of 0 indicates the least accuracy (pants-fire) and a score of 5 indicates the highest accuracy (true).
Consequently, the right term of the loss function in \autoref{eq:loss} aims to minimize the distance between predicted and ground-truth scores, thereby capturing the semantic distance between the predicted and ground-truth labels.

\section{Model properties}
\label{sec:properties}

In this section, we explore the key features of the DHGAT model. This model introduces the concept of "optimal" neighborhoods and dynamic computation graphs, addressing the limitations of standard GNNs. With the ability to learn and optimize the computation graph at each layer, DHGAT not only mitigates issues like over-smoothing and over-squashing but also generates embedding vectors tailored to the specific target task.
\begin{itemize}
	\item
	GNNs generate embeddings through a computational graph defined based on local neighborhoods.
	In this computational graph, node identifiers are not important; only nodes' features matter.
	In other words, if two nodes have computational graphs with the same local structure
	and neighborhood
	and identical node features, the embedding vectors for these two nodes will be the same.
	In DHGAT, each node learns to use an optimal neighborhood, which can lead to the creation of different optimal computational graphs for two nodes with the same neighborhood and features.
	\item
	In standard GNNs, the computational graph of each node is determined based on the initial graph structure and remains fixed throughout the training process.
	However, in DHGAT, each node learns to interact only with optimal neighbors at each layer.
	As a result, the computational graph for each node is dynamic and can vary across different layers, with its optimal structure being learned during the training process.
	\item 
	In GNNs, 
	node embeddings are updated based on their local neighborhoods and are independent of the target task defined in the prediction head. In DHGAT, each node learns an optimal neighborhood for updating its embedding, resulting in a task-specific and optimal computational graph. For instance, if a specific task requires each node to use only a particular type of neighborhood, the DHGAT model can learn to use only that specific type of neighborhood for updating the embeddings.
	\item
	Over-squashing in the context of GNNs refers to a phenomenon where information from distant nodes in a graph gets "squashed" or compressed as it propagates through the network layers.
	This leads to a loss of important details and hampers the network's ability to capture the relationships between distant nodes effectively \cite{topping2022understanding, di2023over}.
	Our model, which allows each node in each layer to choose the optimal type of neighborhood can help in mitigating the over-squashing problem. 
	The optimal selection of information propagation paths and the effective allocation of resources to nodes allow for more efficient information propagation, reducing the problems associated with excessive compression.
	To argue how DHGAT improves over-squashing, we can consider the following points:
	\begin{itemize}
		\item 
		By allowing each node to select the optimal neighborhood type at each layer, DHGAT can optimally receive information from different types of edges during each learning step. This means that important and useful information is transferred more effectively and precisely to the nodes, preventing it from being overly compressed.
		
		\item
		Over-squashing typically occurs on long paths in graphs~\cite{topping2022understanding, di2023over}.
		By choosing the optimal type of neighborhood in DHGAT, nodes can select shorter and more efficient paths for information propagation, which can help reduce the negative effects of over-squashing.
		
		\item
		By allowing each node to learn independently its neighborhood type, the model can be specifically optimized for each node's local structure.
		This specialization can help mitigate issues related to over-squashing.
		
		\item
		There is a formal definition of over-squahsing as the insensitivity of an $L$-layer GNN output at node $v$ to the input features of a distant node $u$, expressed through a bound on the Jacobian \cite{topping2022understanding, DBLP:conf/icml/FinkelshteinHBC24}:	
		\begin{equation}
			\left\Vert \frac{\partial h_v^L}{\partial x_u} \right\Vert \le Q^L {\hat{A}^L}_{vu},
		\end{equation}
		where $Q$ encapsulates architecture-related constants such as network width and activation function smoothness, while the normalized adjacency matrix $\hat{A}$ captures the structural effects of the graph.
		Based on this definition, one of the strategies to reduce over-squashing is to rewire the graph, such that by generating a more optimal structure and adjusting the matrix $\hat{A}$, the upper bound of the above equation increases, thereby reducing the impact of over-squashing \cite{DBLP:conf/icml/FinkelshteinHBC24}.	
		Given that in DHGAT, by optimally selecting the type of neighborhood for each node at each layer, a form of graph rewiring occurs, this network can learn to choose the optimal neighborhood type in a way that information is efficiently propagated from node $u$ to node $v$.
		In other words, DHGAT can learn the optimal graph structure, thereby reducing the impact of over-squashing.
		
		\item
		To conduct a more precise analysis, we use the mutual information theorem.
		The mutual information of two random variables is a measure of the mutual dependence between the two variables, and is defined as follows \cite{cover1999elements}:
		\begin{equation}
			\label{eq:MI}
			I(X;Y) = H(X) - H(X \vert Y),
		\end{equation}
		where $H(X)$ is the marginal entropy of $X$,
		and $H(X \vert Y)$ is the conditional entropy.
		The conditional mutual information is defined as:
		\begin{equation}
			\label{eq:CMI}
			I(X;Y \vert Z) = H(X \vert Z) - H(X \vert Y,z)
		\end{equation}
		where $H(X \vert Z)$ is the conditional entropy of $X$ given $Z$,
		and $H(X \vert Y,z)$ is the conditional entropy of $X$ given both $Y$ and $Z$.		
		If in \autoref{eq:CMI}, $X = h_v^L$, $Y = x_u$ and $Z = N_r(v)$ then $I(X;Y \vert Z)$ represents the degree of dependency/sensitivity of the embedding of $v$ in layer $L$ 
		to the feature vector of $u$, provided that this embedding  was generated using the neighborhood type $r$:
		\begin{equation}
			I(h_v^L;x_u \vert N_r(v)) =
			H(h_v^L \vert N_r(v)) - H(h_v^L \vert x_u,N_r(v))
		\end{equation}
		Therefore, an increase in $I(h_v^L;x_u \vert N_r(v))$ signifies an increase in the mutual information of the embedding vector of node $v$ and the feature vector of node $u$, which could be a distant node. 
		This increase leads to a reduction in over-squashing.
		
		We demonstrate that DHGAT can increase $I(h_v^L;x_u \vert N_r(v))$ by optimally selecting the neighborhood type.
		In GNNs, the neighborhood $N_r(v)$ acts as an information filter.
		The choice of $r$ determines which information from neighboring nodes contributes to the embedding $h_v^L$.
		Let $r^*$ be the optimal filter, meaning it allows the most relevant information to pass through.
		The optimal neighborhood $N_{r^*}(v)$  is chosen such that it maximizes the relevant information passed to $h_v^L$, thereby increasing $I(h_v^L;x_u \vert N_r(v))$.
		Formally:
		\begin{equation}
			\label{eq:HX-Z}
			\forall r \neq r^* : H(h_v^L \vert N_{r^*}(v)) \leq H(h_v^L \vert N_r(v)).
		\end{equation}
		This implies that the entropy of $h_v^L$ given the optimal neighborhood is lower,
		meaning $h_v^L$ is more informative or predictable given $N_{r^*}(v)$.
		When considering both $x_u$ and $N_r(v)$, the conditional entropy $H(h_v^L \vert x_u,N_r(v))$ captures the remaining uncertainty about $h_v^L$,
		given the input features of $u$ and the neighborhood of $v$.
		The optimal choice of $N_{r^*}(v)$ is expected to minimize this entropy because it preserves the most relevant information from $x_u$:
		\begin{equation}
			\label{eq:HX-YZ}
			\forall r \neq r^* :
			H(h_v^L \vert x_u , N_{r^*}(v)) \leq H(h_v^L \vert  x_u  , N_r(v)).
		\end{equation}
		
		Given the results from \autoref{eq:HX-Z} and \autoref{eq:HX-YZ}, we have:
		\begin{equation}
			\forall r \neq r^* :
			I(h_v^L;x_u \vert N_{r^*}(v)) \geq I(h_v^L;x_u \vert N_r(v)).
		\end{equation}
		This inequality shows that the conditional mutual information is maximized when $N_{r^*}(v)$ is selected, meaning the optimal neighborhood selection maximizes the information shared between the node embedding $h_v^L$ and the input feature $x_u$.
		This maximization is due to the optimal neighborhood's ability to effectively filter and pass relevant information, reducing uncertainty and enhancing the embedding quality, helping to mitigate over-squashing.		
	\end{itemize}
	
	\item	
	Over-smoothing is a phenomenon in GNNs where, as the number of layers in the network increases, the node embeddings tend to become increasingly similar \cite{li2018deeper}.
	The ability of DHGAT to allow nodes to independently choose their optimal neighborhood type can potentially mitigate over-smoothing.
	\begin{itemize}
		\item 
		Traditional GNNs apply a uniform aggregation of neighbor embeddings across all nodes, which can lead to rapid information propagation and over-smoothing.
		However, in our model, each node independently selects the most relevant neighbors (based on edge types) for aggregation.
		This selective aggregation implies that information propagation is more controlled and customized.
		By preventing all nodes from using the same aggregation strategy, the model reduces the risk of all node features converging to similar values, which is a direct cause of over-smoothing.
		
		\item
		The independent and optimal learning of the neighborhood type for each node allows the model to better preserve the unique features of each node and prevent excessive feature convergence.
		This ensures that each node retains its specific information more effectively, thereby reducing the risk of over-smoothing.
		
	\end{itemize}
	
\end{itemize}
\section{Experiments}
\label{sec:exp}
In this section, first we provide an overview of the LIAR dataset, the baseline models and the evaluation metrics used in our empirical evaluations. 
Then, we report the results of our extensive experiments, encompassing diverse settings and conditions to comprehensively assess our model \footnote{The code of our method is publicly available at: \url{https://github.com/blakzaei/DHGAT}.}.

\subsection{Dataset}
\label{sec:ds}

We evaluate the performance of our proposed model on the LIAR~\cite{DBLP:conf/acl/Wang17} dataset, one of the largest public benchmarks for real-world fake news detection.
The dataset comprises 12,836 short labeled news from 3,318 public speakers, with an average of 17.9 tokens per news item.
These news are labeled into six categories: pants-fire, false, barely-true, half-true, mostly-true, and true.
The LIAR dataset is created in 2017 and sourced from PolitiFact's public fake news data.
It includes extensive speaker profiles, such as speaker name, party affiliations, job title, home state, context, subject, and credit history.
\autoref{tab:ds_example} illustrates two examples from this dataset.

\begin{table}[]
	\centering
	\caption{Two examples from the LIAR dataset.\label{tab:ds_example}}
	\resizebox{\textwidth}{!}{%
		\begin{tabular}{cll}
			\hline
			\multicolumn{1}{l}{}        & \multicolumn{1}{c}{\textbf{example 1}} & \multicolumn{1}{c}{\textbf{example 2}} \\ \hline
			\textbf{statement} &
			\makecell[l]{Twelve judges have thrown out legal\\ challenges to the health care law because they \\ rejected the notion that the health care\\ law was unconstitutional.} &
			\makecell[l]{Gov. Bob McDonnellsbudget plantakesmoney  \\out of our classrooms to pave roads.} \\ \hline
			\textbf{speaker}            & barack-obama                           & donald-mceachin                        \\ \hline
			\textbf{subject}            & health-care,supreme-court              & education,state-budget                 \\ \hline
			\textbf{job\_title}         & President                              & State senator                          \\ \hline
			\textbf{state\_info}        & Illinois                               & Virginia                               \\ \hline
			\textbf{party\_affiliation} & democrat                               & democrat                               \\ \hline
			\textbf{context}            & an interview on Fox                    & a speech                               \\ \hline
			\textbf{credit-history}     & (70, 71, 160, 163, 9)                  & (0, 1, 0, 0, 0)                        \\ \hline
			\textbf{label}              & 0: False                               & 3: True                                \\ \hline
		\end{tabular}%
	}
\end{table}

\subsection{Baseline Models}
\label{sec:base_lines}
To evaluate the performance of multi-class fake news detection on the LIAR dataset, we compare the DHGAT model against seven fake news detection methods:
\begin{itemize}
	\item \textbf{LSTM-CP-GCN} \cite{ren2023fake} explores intra-article interactions by representing each article as a weighted graph, with sentences as vertices. An LSTM computes feature vectors for the vertices, while a CP decomposition-based method calculates the weight matrix using local word co-occurrence information of sentences. The graph is then fed into a GCN for classification.
	
	\item \textbf{HGNNR4FD} \cite{xie2023heterogeneous} constructs a heterogeneous graph using news content and incorporates Knowledge Graph (KG) relations to explore ground-truth knowledge. It employs an attention-based heterogeneous graph neural network to aggregate information from both the heterogeneous graph and KG, enhancing the learned news representations for classification.
	
	\item \textbf{MGCN} \cite{hu2019multi} utilizes multi-depth GCN blocks to capture multi-scale information from node neighbors, integrating this information through an attention mechanism.
	
	\item \textbf{TextGCN} \cite{yao2019graph} represents the entire corpus as a heterogeneous graph consisting of word and news nodes. It uses a GCN to learn embeddings for both types of nodes, effectively capturing the relationships between words and news items to classify the news nodes.
	
	\item \textbf{TextCNN} \cite{Chen2015} applies CNNs to classify news text.
	
	\item \textbf{GATv2} \cite{DBLP:conf/iclr/Brody0Y22} leverages speaker profile information to construct a graph, and then uses standard GATv2 to learn embeddings for news classification.
	
	\item \textbf{GCN} \cite{DBLP:conf/iclr/KipfW17}  utilizes speaker profile information to construct a graph, and then employs standard GCN to learn embeddings for news classification.
\end{itemize}

\subsection{Experimental setup}
\label{sec:setup}

We use the pre-trained FastText model to extract the feature vectors of the textual content of news articles. The 300-dimensional embedding vectors are considered as the textual feature vectors for each news item. To construct a heterogeneous graph, we consider various types of neighborhoods.
In the LIAR dataset, we define 9 types of edges: speaker, context, subject, party-affiliation, job-title, state, KNN-5, KNN-6, and KNN-7. The first six are side information available in the dataset, and the last three are based on the similarity of textual embedding vectors using the K-nearest neighbors (KNN) algorithm. We explore different combinations of subsets (pairs, triplets, etc.) from these edge types to construct the heterogeneous graph.
For both the decision network ($\Phi$) and representation network ($\Psi$), we use a 2-layer GATv2 network. The hidden units are set to $[256, 128]$, and the learning rate is $0.001$. The dropout rate is set to $0.5$. We use the Adam optimizer with a weight decay strategy to train all parameters over $200$ epochs.
Since the LIAR dataset is fairly balanced, we, like most baseline methods and current approaches, use accuracy and F1-score as the primary evaluation metrics.

Our proposed approach is semi-supervised. Therefore, we consider three different scenarios for the proportion of labeled data: 10\%, 20\%, and 30\%. Each model is tested 10 times through random splits of the training and test sets, and we report the mean and standard deviation of the experimental results. Unless explicitly noted, all baseline parameters are set the same as DHGAT to ensure fairness. All of the mentioned methods are implemented in Python using PyTorch libraries.

\subsection{Results}
\label{sec:results}

\begin{table}[]
	\centering
	\caption{Performance of fake news detection models on the LIAR dataset.\label{tab:final_results}}
	\resizebox{\textwidth}{!}{%
		\begin{tabular}{ccccccc}
			\hline
			\multirow{2}{*}{\textbf{method}} & \multicolumn{2}{c}{\textbf{10\%}} & \multicolumn{2}{c}{\textbf{20\%}} & \multicolumn{2}{c}{\textbf{30\%}} \\ 
			& \textbf{ACC}  & \textbf{F1-score} & \textbf{ACC}  & \textbf{F1-score} & \textbf{ACC}  & \textbf{F1-score} \\ \hline
			DHGAT   & 0.4568$\pm$0.0109 & 0.4042$\pm$0.0156     & 0.4848$\pm$0.015  & 0.4376$\pm$0.0351     & 0.5051$\pm$0.0118 & 0.4763$\pm$0.0114     \\ \hline
			LSTM-CP-GCN & 0.2984$\pm$0.0035 & 0.1872$\pm$0.0291     & 0.3043$\pm$0.0022 & 0.1630$\pm$0.0102      & 0.3003$\pm$0.0051 & 0.1573$\pm$0.0019     \\ \hline
			HGNNR4D & 0.1992$\pm$0.0078 & 0.113$\pm$0.0153      & 0.2137$\pm$0.0085 & 0.1137$\pm$0.0125     & 0.2029$\pm$0.0081 & 0.0802$\pm$0.011      \\ \hline
			MGCN    & 0.4252$\pm$0.0068 & 0.3925$\pm$0.0089     & 0.4586$\pm$0.0075 & 0.4359$\pm$0.0095     & 0.4654$\pm$0.0035 & 0.4398$\pm$0.0085     \\ \hline
			TextGCN & 0.3176$\pm$0.0054 & 0.2645$\pm$0.0105     & 0.319$\pm$0.0041  & 0.2701$\pm$0.0092     & 0.3181$\pm$0.0103 & 0.2678$\pm$0.0064     \\ \hline
			TextCNN & 0.2996$\pm$0.0034 & 0.208$\pm$0.0148      & 0.3056$\pm$0.0062 & 0.2158$\pm$0.0149     & 0.2996$\pm$0.0054 & 0.1809$\pm$0.0118     \\ \hline
			GAT     & 0.3904$\pm$0.0114 & 0.3251$\pm$0.021      & 0.4202$\pm$0.0121 & 0.362$\pm$0.0254      & 0.4419$\pm$0.0155 & 0.4217$\pm$0.0202     \\ \hline
			GCN     & 0.3884$\pm$0.0104 & 0.3335$\pm$0.0248     & 0.4203$\pm$0.0137 & 0.3791$\pm$0.0305     & 0.4308$\pm$0.0095 & 0.3933$\pm$0.0154     \\ \hline
		\end{tabular}%
	}
\end{table}

\autoref{tab:final_results} shows the results for 6-class fake news classification on the LIAR dataset for various models. Our proposed DHGAT model, along with baseline models like GATv2, GCN, and MGCN, utilizes different types of contextual information to construct the graph. We have reported the best result obtained for each model in \autoref{tab:final_results}.
The TextCNN model, which only uses the textual content of news for classification, generally performs worse than the graph-based models. This suggests the positive impact of modeling the dataset as a graph and learning the structural relationships between different data points.
The HGNNR4D model is originally proposed for binary classification of fake news on the LIAR dataset. We adapted it with slight modifications to its classification layer for multi-class classification. Despite achieving relatively good performance in binary classification on the LIAR dataset
(accuracy of 57.53, as stated by the original authors), it shows very poor performance in the 6-class classification task.
The LSTM-CP-GCN and TextGCN models use a graph-based modeling approach. However, the textual content of each news item is modeled separately and independently as an individual graph, and the task is defined as a graph classification problem. In these models, no relationships between different news items are considered, and each news article is processed independently.
Finally, the GAT, GCN, and MGCN models utilize contextual information to model the dataset as a graph, where each news item is treated as a node and the edges represent relationships between different data points. These models generally outperform the other evaluated methods. GATv2 performs better than GCN because, unlike GCN, which treats the influence of all neighboring nodes equally, GATv2 assigns different weights to each neighbor, and these weights are learned during the training process. None of these models have a control mechanism to select optimal neighborhood connections.
The DHGAT model, with its ability to learn and select the optimal neighborhood for each node, achieves better performance than the aforementioned methods, improving accuracy by approximately 4\%.

\begin{figure}[!t]
	\centering
	\subfigure[]{\includegraphics[width=0.48\textwidth]{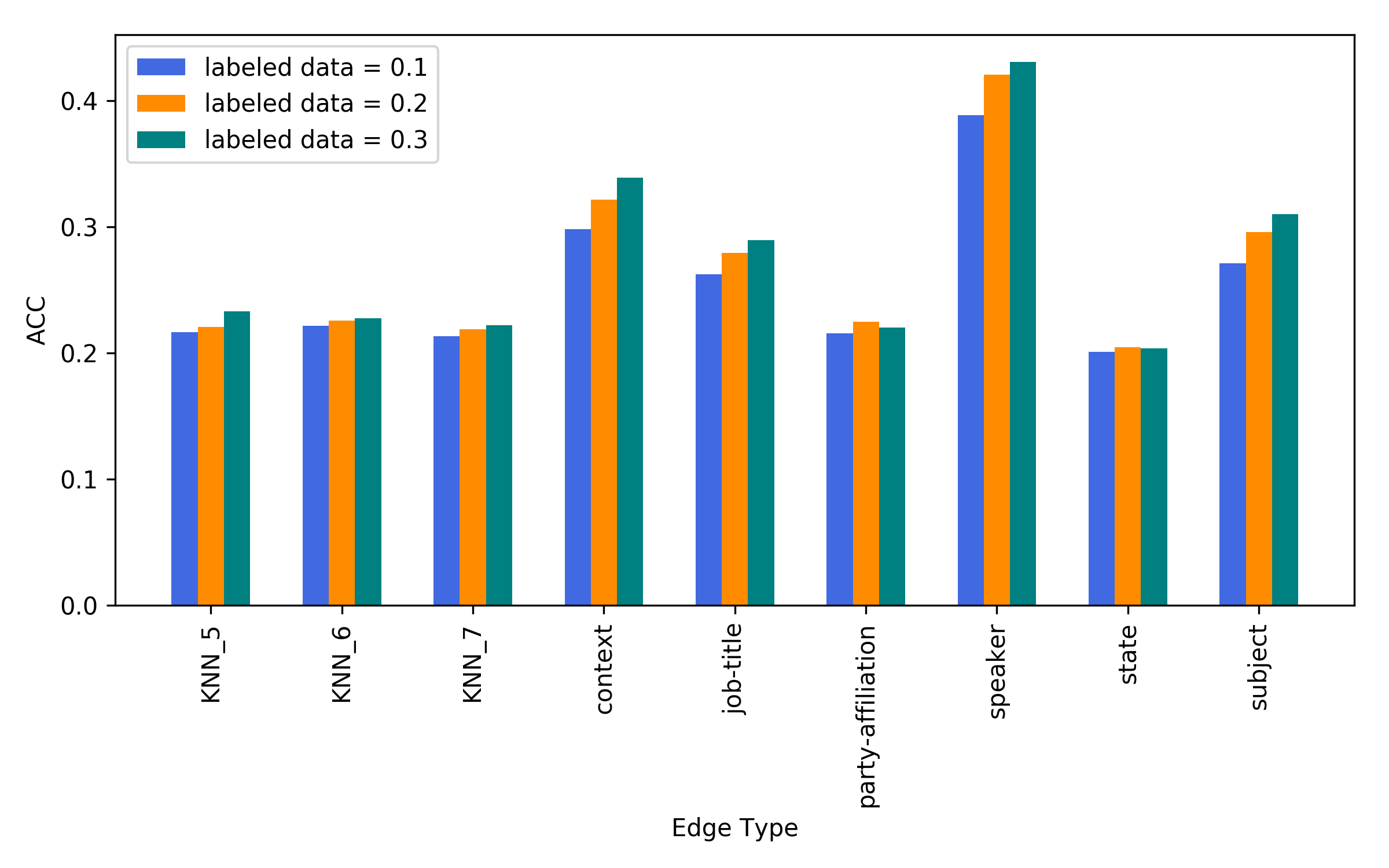}}
	\subfigure[]{\includegraphics[width=0.48\textwidth]{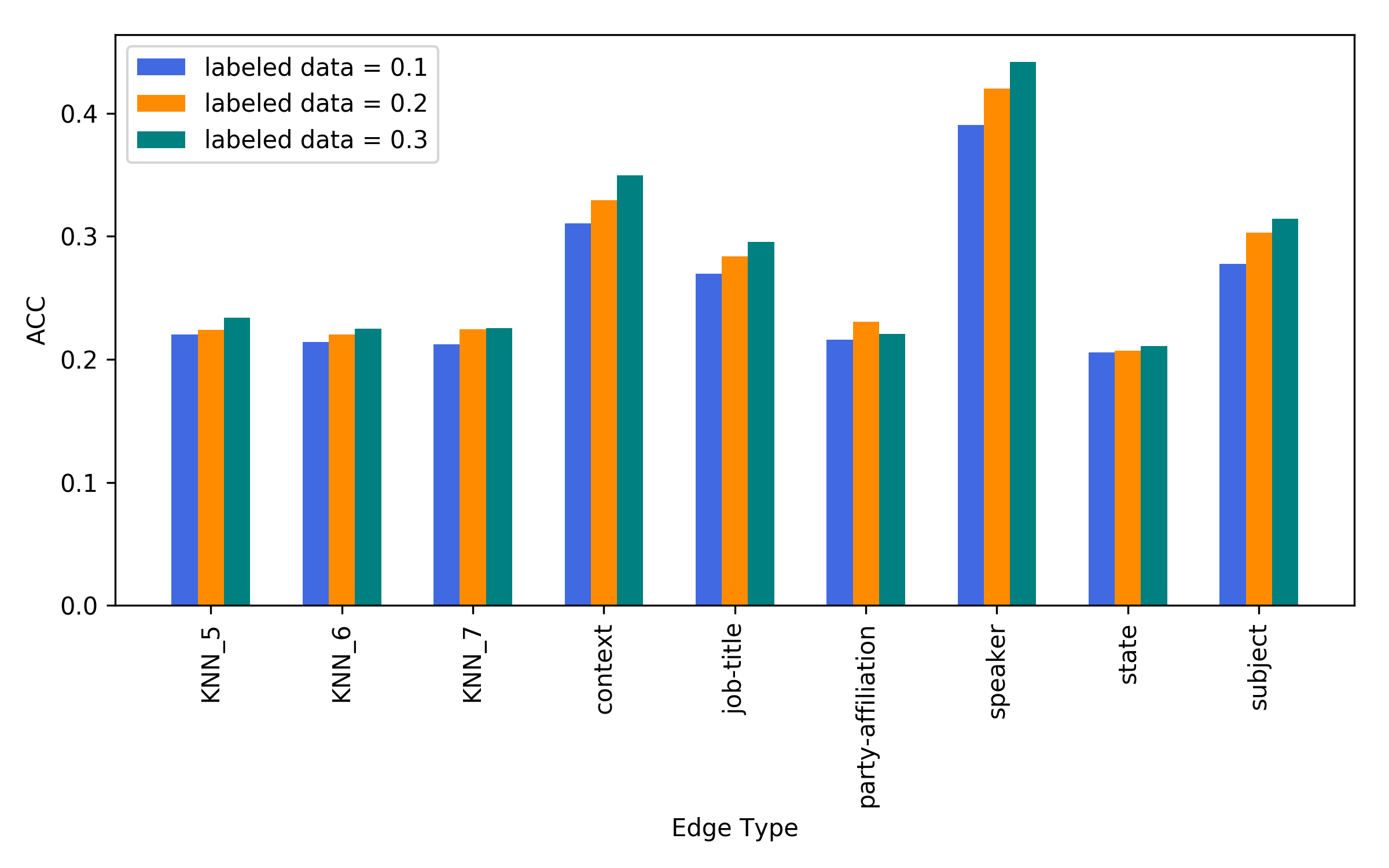}}
	\caption{Accuracy of fake news detection using the models: a) GCN and b) GAT,
		on the graphs constructed based on various speaker profile attributes and the similarity of textual embedding vectors.\label{fig:edge_type}}
\end{figure}

To understand the impact of each speaker profile attribute (speaker, context, subject, party affiliation, job title, state) and the textual embedding vectors (KNN-5, KNN-6, KNN-7), we present the results of graph construction based on each of these features. These graphs are then processed using the GATv2 and GCN models, as shown in \autoref{fig:edge_type}.
As expected, constructing the graph using the speaker feature delivers the best performance, consistent with reports from existing methods. Following that, the context and subject features rank second and third, respectively.
Using the state and party affiliation features for graph construction does not model the relationships between news data as effectively. In some cases, even using textual embedding similarities (KNN-5, KNN-6, and KNN-7)--although with slight differences--yields better graphs.

\begin{figure}[!t]
	\centering		
	\includegraphics[scale=.5]{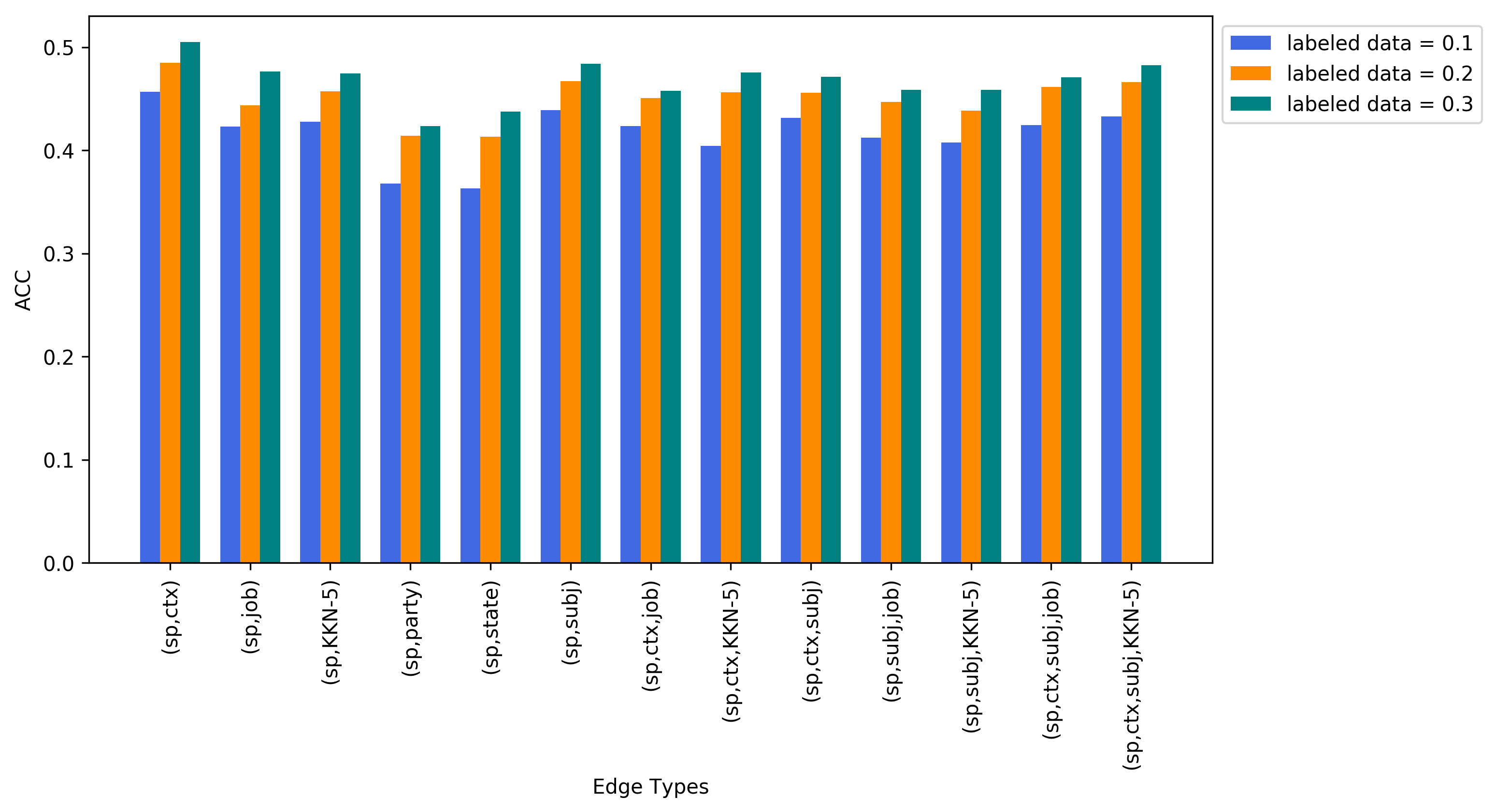}
	\caption{Accuracy of fake news detection using the DHGAT model on heterogeneous graphs constructed with various edge types. For brevity, the edge types in the figure are summarized as follows: sp: speaker, ctx: context, job: job-title, subj: subject.\label{fig:dhgat_graph_size}}
\end{figure}

For constructing the heterogeneous graph in the DHGAT model, we considered various subsets of the edge types mentioned above. The accuracy achieved based on the chosen edge types for graph construction is shown in \autoref{fig:dhgat_graph_size}. As depicted in this figure, the highest accuracy in fake news detection is obtained when the heterogeneous graph is constructed using the context and speaker edge types.
Comparing this figure with \autoref{fig:edge_type} reveals that, in general, combining edge types that individually produce better graphs results in a superior heterogeneous graph. For instance, \autoref{fig:edge_type} shows that using the speaker and context edge types yields the highest accuracy. Similarly, the heterogeneous graph constructed using these two edge types also achieves the highest accuracy.

\begin{figure}[!t]
	\centering		
	\includegraphics[scale=.5]{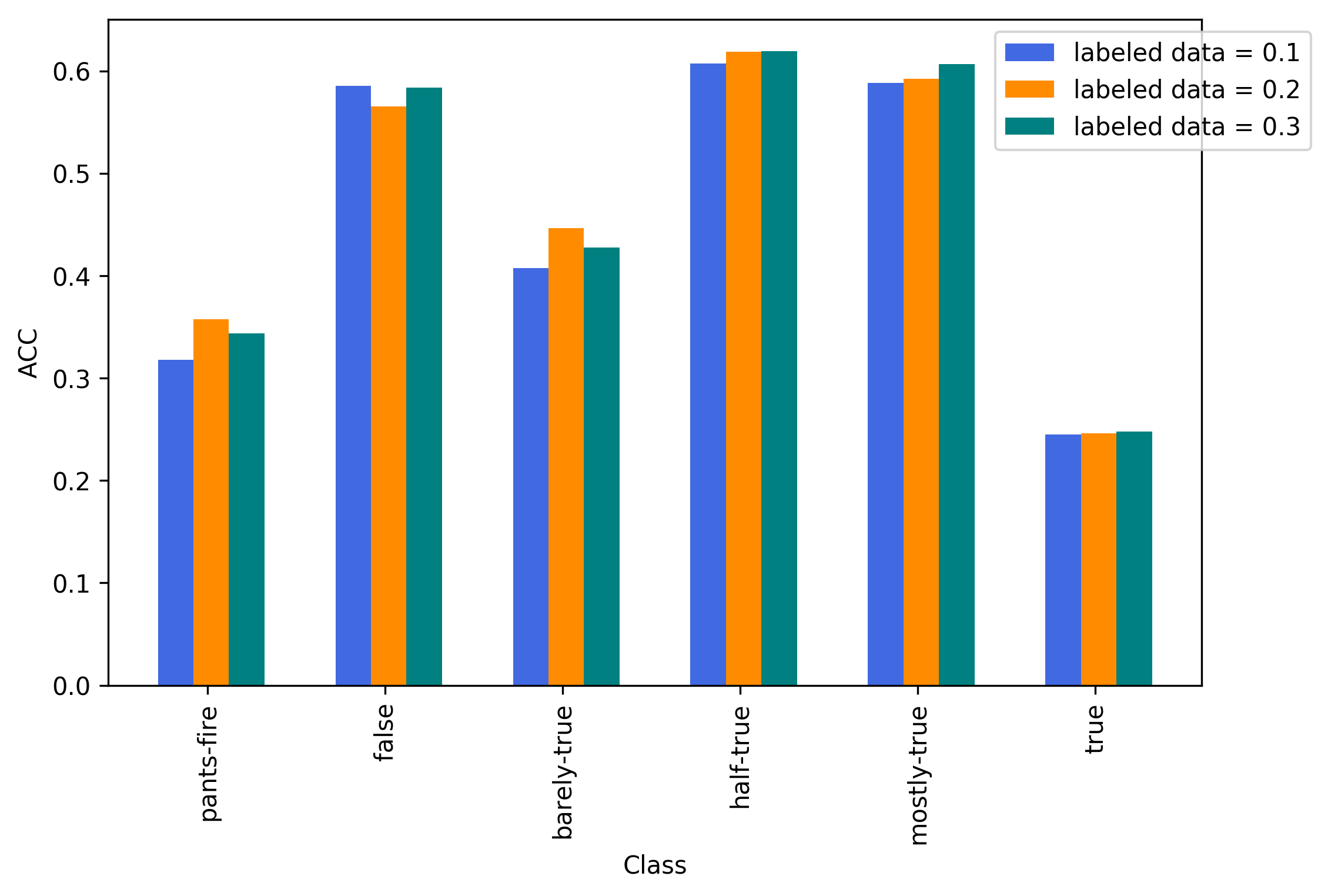}
	\caption{Accuracy of fake news detection for each class by the DHGAT model on the heterogeneous graph constructed with two types of edges: speaker and context.\label{fig:class_acc}}
\end{figure}

\begin{figure}[!t]
	\centering
	\subfigure[]{\includegraphics[width=0.3\textwidth]{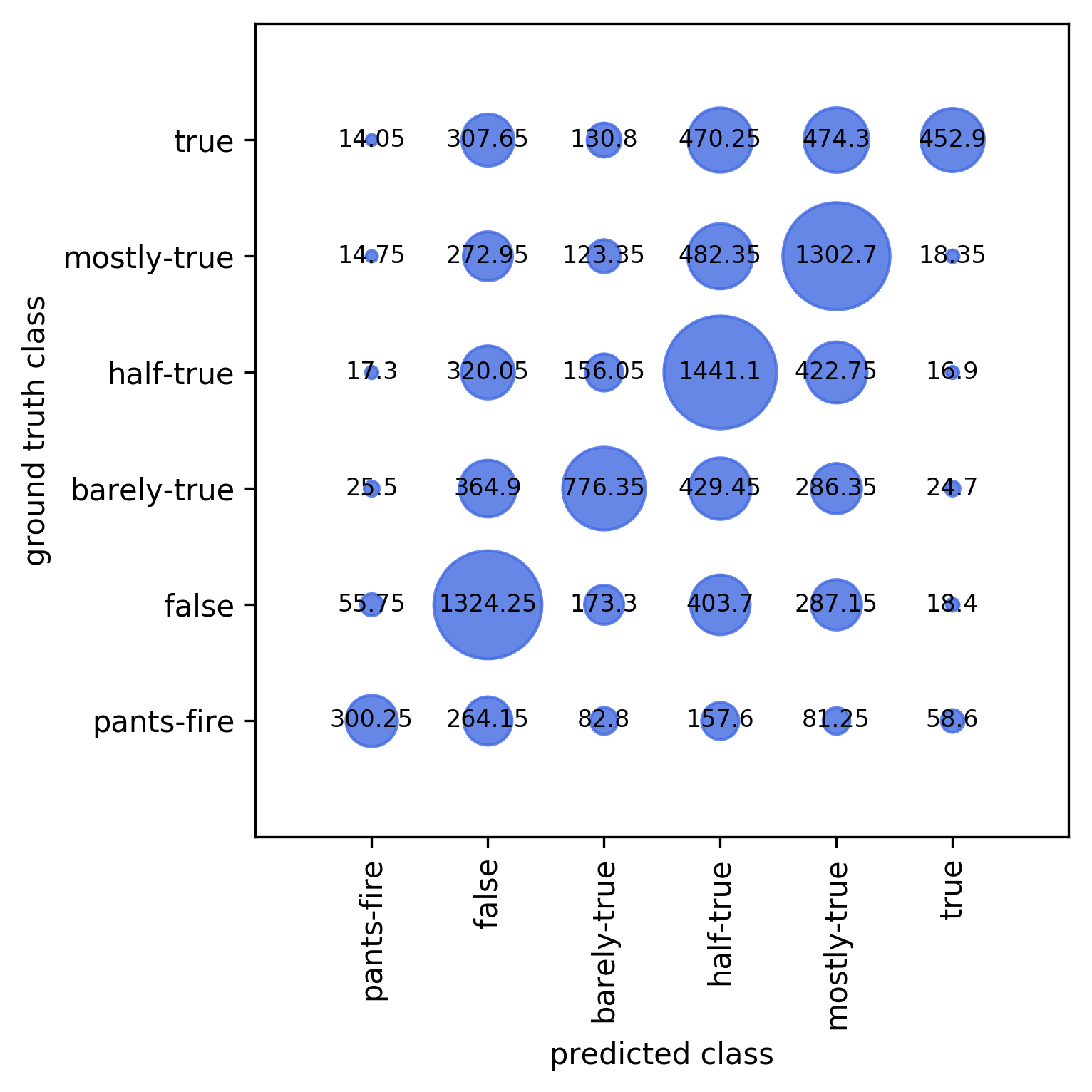}}
	\subfigure[]{\includegraphics[width=0.3\textwidth]{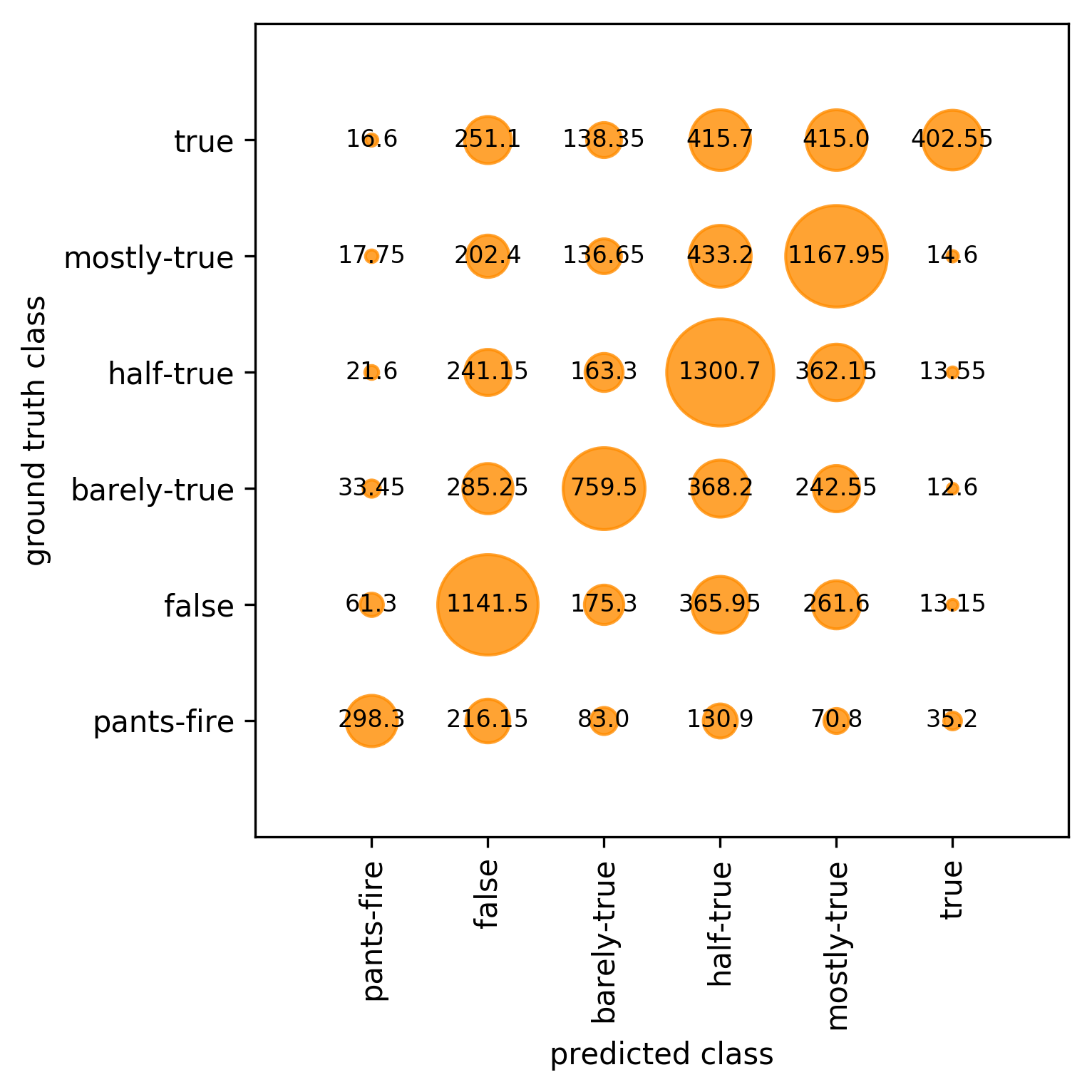}}
	\subfigure[]{\includegraphics[width=0.3\textwidth]{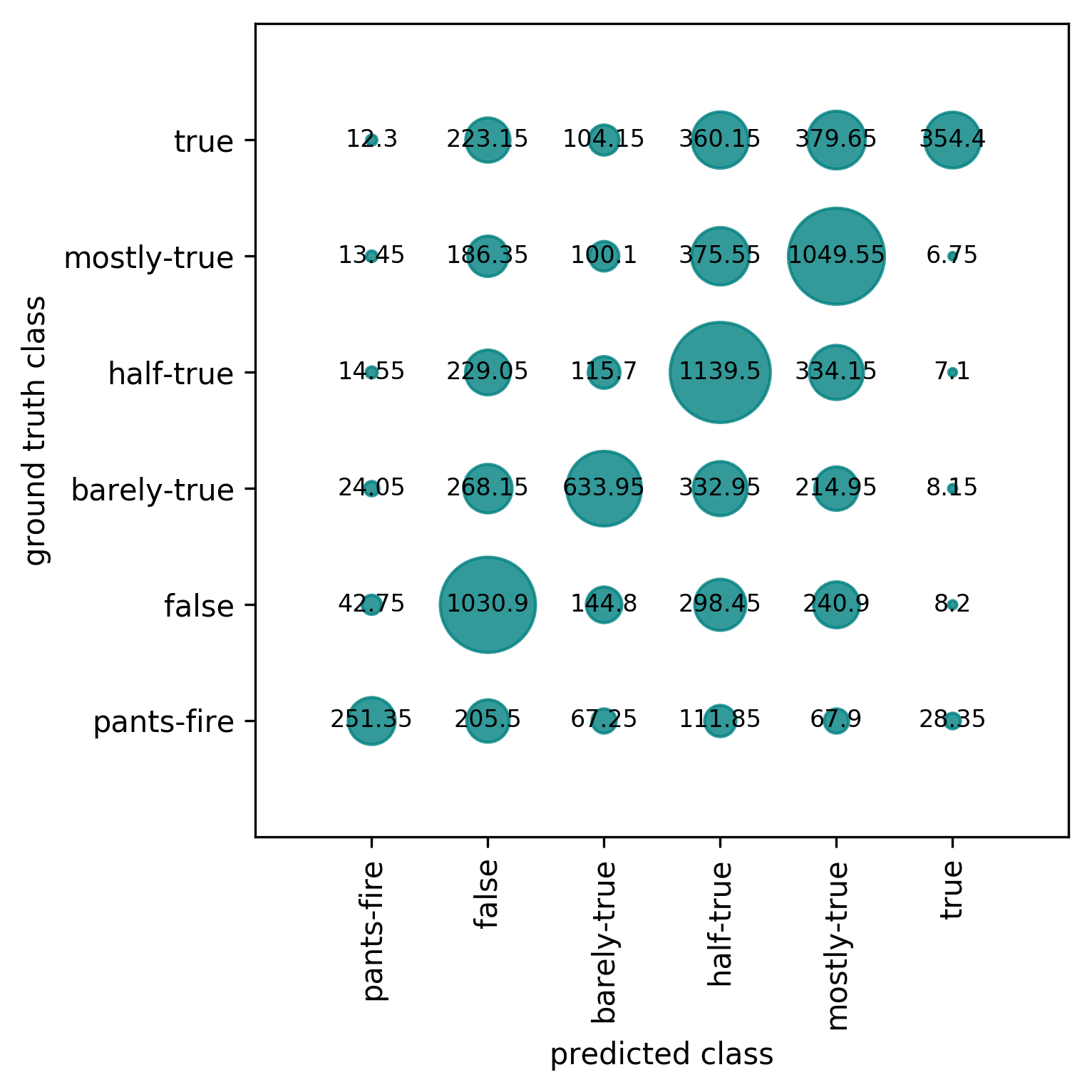}}
	\caption{Confusion matrix for the average of ten runs of the DHGAT model on unlabeled data for three labeling scenarios: a) 10\%, b) 20\%, and c) 30\%.\label{fig:cm}}
\end{figure}

To assess the performance of the DHGAT model on each class, we plot both the accuracy and the confusion matrix, averaged over ten runs. The model's evaluation is conducted on the unlabeled data using the heterogeneous speaker-context graph, which demonstrates the highest performance as indicated in \autoref{fig:dhgat_graph_size}. The detailed class-wise accuracy is shown in \autoref{fig:class_acc}, while the confusion matrix highlighting misclassifications is presented in \autoref{fig:cm}.

\autoref{fig:class_acc} illustrates that the model excels in classifying the "false," "half-true," and "mostly-true" categories but shows weaker performance in the "true" and "pants-fire" categories. Overall, as the amount of labeled data increases, accuracy generally improves across most classes, though this enhancement is not uniform. Certain classes, such as "pants-fire" and "barely-true," are more sensitive to the quantity of labeled data, highlighting the greater complexities associated with these categories.

\autoref{fig:cm} presents the confusion matrix for the average of ten runs on the unlabeled data. To enhance visualization, we adjust the circle diameters to reflect the numerical values. A key observation is that when labels are incorrectly predicted (represented by circles outside the diagonal), the error rate typically decreases as the semantic distance between the actual and predicted labels increases. For instance, errors in the "pants-fire" class are most commonly predicted as "false," which has the smallest semantic distance to "pants-fire," while predictions as "true," which has the largest semantic distance, are less frequent. This pattern is consistent across other classes as well. This indicates that the second term in the loss function (\autoref{eq:loss}) significantly helps in minimizing semantic errors, as incorrect labels are often close in semantic meaning to the correct ones. Furthermore, the confusion matrix reveals that the number of misclassified samples in the "false" and "half-true" classes is higher compared to other classes, possibly due to less distinctive features for these categories.

\begin{figure}[!t]
	\centering
	\subfigure[]{\includegraphics[width=0.7\textwidth]{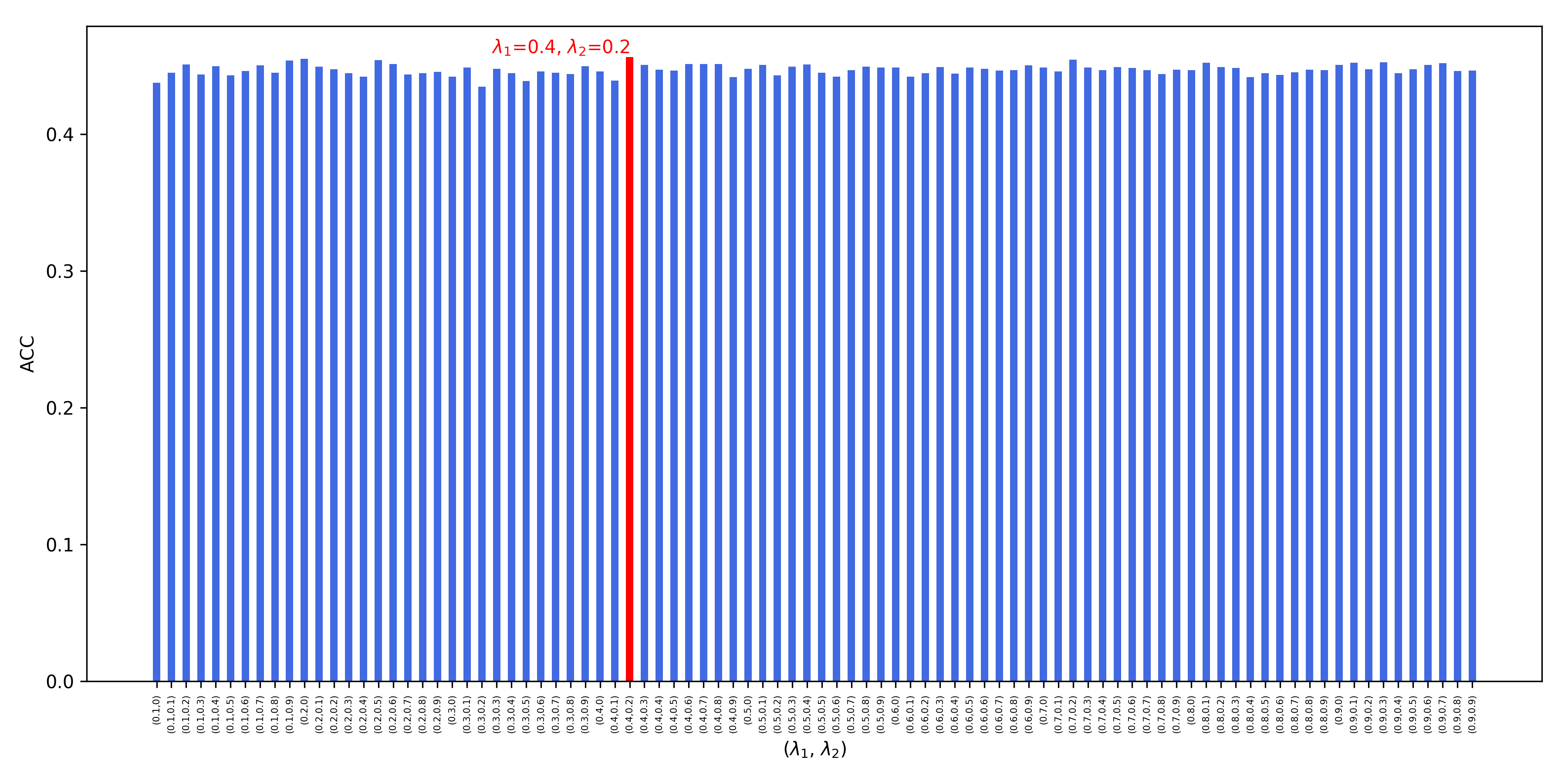}}
	\subfigure[]{\includegraphics[width=0.7\textwidth]{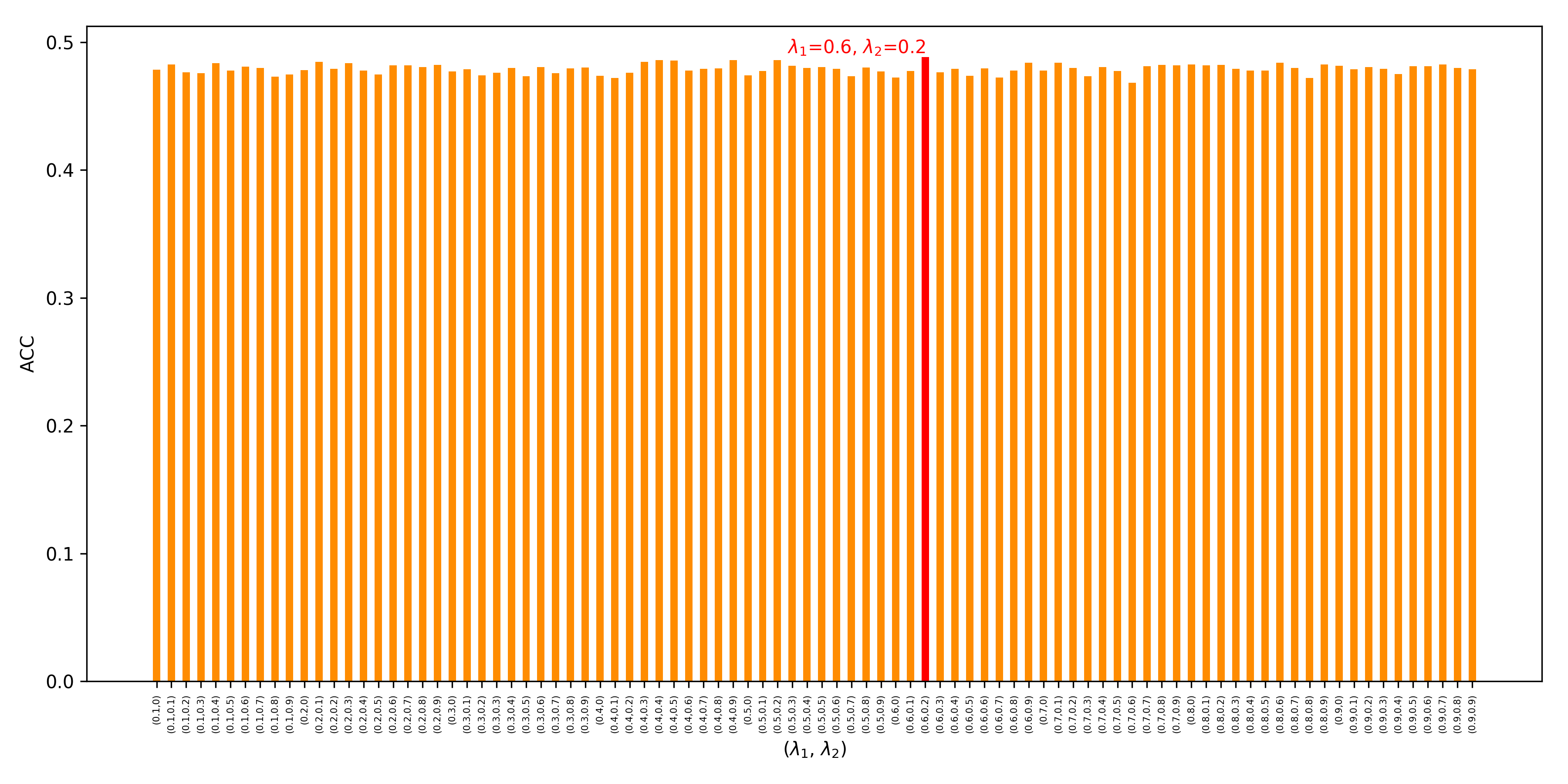}}
	\subfigure[]{\includegraphics[width=0.7\textwidth]{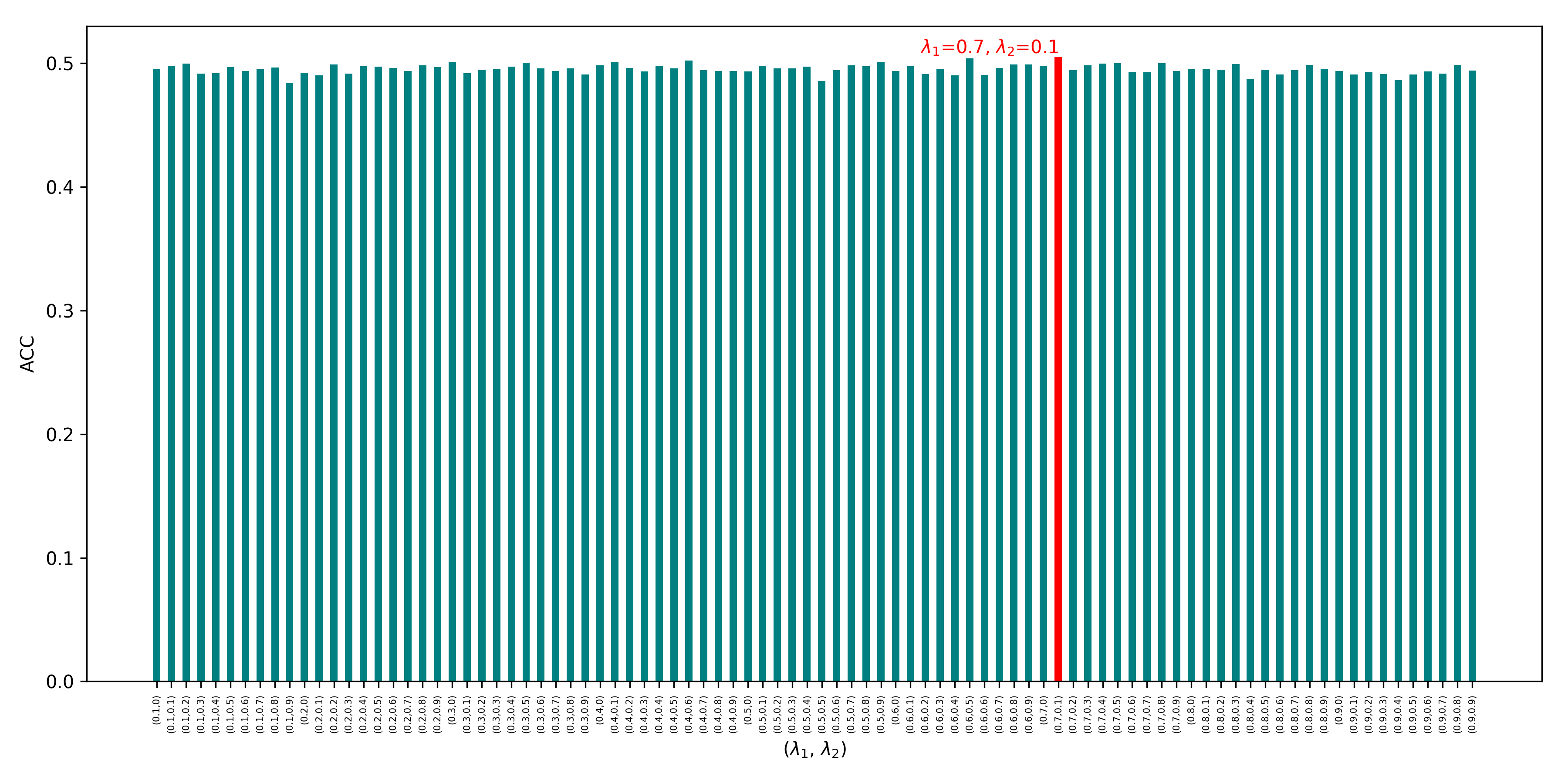}}
	\caption{Variation in model accuracy based on different settings of the parameters $\lambda_1$ and $\lambda_2$, for three labeling scenarios: a) 10\%,  b) 20\%, and c) 30\%.\label{fig:regularization}}
\end{figure}

To conduct a more detailed analysis of the impact of the two different components of the loss function (\autoref{eq:loss}), we plot the average model accuracy for the heterogeneous graph constructed from two types of edges: speaker and context. This is done with respect to various values of the $\lambda_1$ and $\lambda_2$ 
regularization parameters, as shown in \autoref{fig:regularization}. The horizontal axis represents different values set for $\lambda_1$ and $\lambda_2$, while the vertical axis shows the average accuracy over ten runs of model evaluation on the unlabeled data. In each chart, the values that result in the highest accuracy are highlighted in red.
The results show that, in general, the left term of the loss function, which calculates the classification error based on cross-entropy, exerts more influence and requires a higher weight compared to the right term, which aims to reduce the semantic distance between the true labels and the predicted labels.
Despite this, the model demonstrates improved performance when the right term is included in the cost function, compared to when the cost function is calculated based solely on cross-entropy (i.e., when $\lambda_2 = 0$).

\section{Conclusion}
\label{sec:conclusion}

In this paper, we introduced the Decision-based Heterogeneous Graph Attention Network (DHGAT), a novel semi-supervised model for multi-class fake news detection. DHGAT effectively addresses several limitations inherent in traditional graph neural networks by dynamically optimizing neighborhood types at each layer for every node. It represents news data as a heterogeneous graph where nodes (news items) are connected through multiple types of edges that capture various relationships.
The DHGAT architecture, comprising a decision network and a representation network, allows nodes to independently choose their optimal neighborhood type in each layer. This dynamic and task-specific approach to neighborhood selection significantly enhances the model's ability to generate accurate and informative embeddings, tailored to the classification task.
Our extensive experiments on the LIAR dataset demonstrated the high accuracy and performance of DHGAT.


\bibliographystyle{unsrt} 
\bibliography{references}

\end{document}